\documentclass{article}

\usepackage{microtype}
\usepackage{graphicx}
\usepackage{subfigure}
\usepackage{booktabs} % for professional tables

\usepackage{hyperref}

\usepackage[accepted]{icml2023}

\usepackage{amsmath}
\usepackage{amssymb}
\usepackage{mathtools}
\usepackage{amsthm}

\usepackage[capitalize,noabbrev]{cleveref}

\theoremstyle{plain}

\theoremstyle{definition}

\theoremstyle{remark}

\usepackage[textsize=tiny]{todonotes}

\usepackage{multirow}
\usepackage{kotex}
\usepackage{wrapfig,lipsum}
\icmltitlerunning{Probabilistic Concept Bottleneck Models}

\begin{document}

\twocolumn[
\icmltitle{Probabilistic Concept Bottleneck Models}



\icmlsetsymbol{equal}{*}

\begin{icmlauthorlist}
\icmlauthor{Eunji Kim}{snuece}
\icmlauthor{Dahuin Jung}{snuece}
\icmlauthor{Sangha Park}{snuece}
\icmlauthor{Siwon Kim}{snuece}
\icmlauthor{Sungroh Yoon}{snuece,snuothers}
\end{icmlauthorlist}

\icmlaffiliation{snuece}{Department of Electrical and Computer Engineering, Seoul National University, Seoul, Korea}
\icmlaffiliation{snuothers}{Interdisciplinary Program in Artificial Intelligence, Seoul National University, Seoul, Korea}

\icmlcorrespondingauthor{Sungroh Yoon}{sryoon@snu.ac.kr}

\icmlkeywords{Machine Learning, ICML, probabilistic embeddings, concept bottleneck models, interpretability}

\vskip 0.3in
]



\printAffiliationsAndNotice{}  % leave blank if no need to mention equal contribution

\newcommand{\eg}{\textit{e.g.,~}}
\newcommand{\ie}{\textit{i.e.,~}}

\begin{abstract}
Interpretable models are designed to make decisions in a human-interpretable manner. Representatively, Concept Bottleneck Models (CBM) follow a two-step process of concept prediction and class prediction based on the predicted concepts. CBM provides explanations with high-level concepts derived from concept predictions; thus, reliable concept predictions are important for trustworthiness. In this study, we address the ambiguity issue that can harm reliability. While the existence of a concept can often be ambiguous in the data, CBM predicts concepts deterministically without considering this ambiguity. To provide a reliable interpretation against this ambiguity, we propose Probabilistic Concept Bottleneck Models (ProbCBM). By leveraging probabilistic concept embeddings, ProbCBM models uncertainty in concept prediction and provides explanations based on the concept and its corresponding uncertainty. This uncertainty enhances the reliability of the explanations. Furthermore, as class uncertainty is derived from concept uncertainty in ProbCBM, we can explain class uncertainty by means of concept uncertainty. Code is publicly available at \url{https://github.com/ejkim47/prob-cbm}.
\end{abstract}

\section{Introduction}\label{sec:intro}

As deep learning systems have been increasingly used in various applications and fields, ensuring transparency of the systems' decision-making has become a significant challenge~\cite{esteva2019guide,miller2019explanation}. Numerous post-hoc explanation methods have been introduced to explain the decision-making of already-trained deep neural networks~\cite{simonyan2013deep,kim2018interpretability,goyal2019counterfactual}. However, post-hoc methods cannot entirely explain the model's prediction~\cite{zhou2018interpretable} and provide approximate explanations in a human-understandable form, which may lead to incorrect explanations~\cite{rudin2019stop}. In contrast, interpretable models are designed to make decisions through human-interpretable processes, ensuring interpretability and not requiring an external explanation method to account for their decisions. Accordingly, research on building interpretable models has been actively conducted~\cite{melis2018towards, chen2019looks, koh2020concept, jung2020icaps, bohle2021convolutional}.

\begin{figure}[t]
\begin{center}
\centerline{\includegraphics[width=0.98\columnwidth]{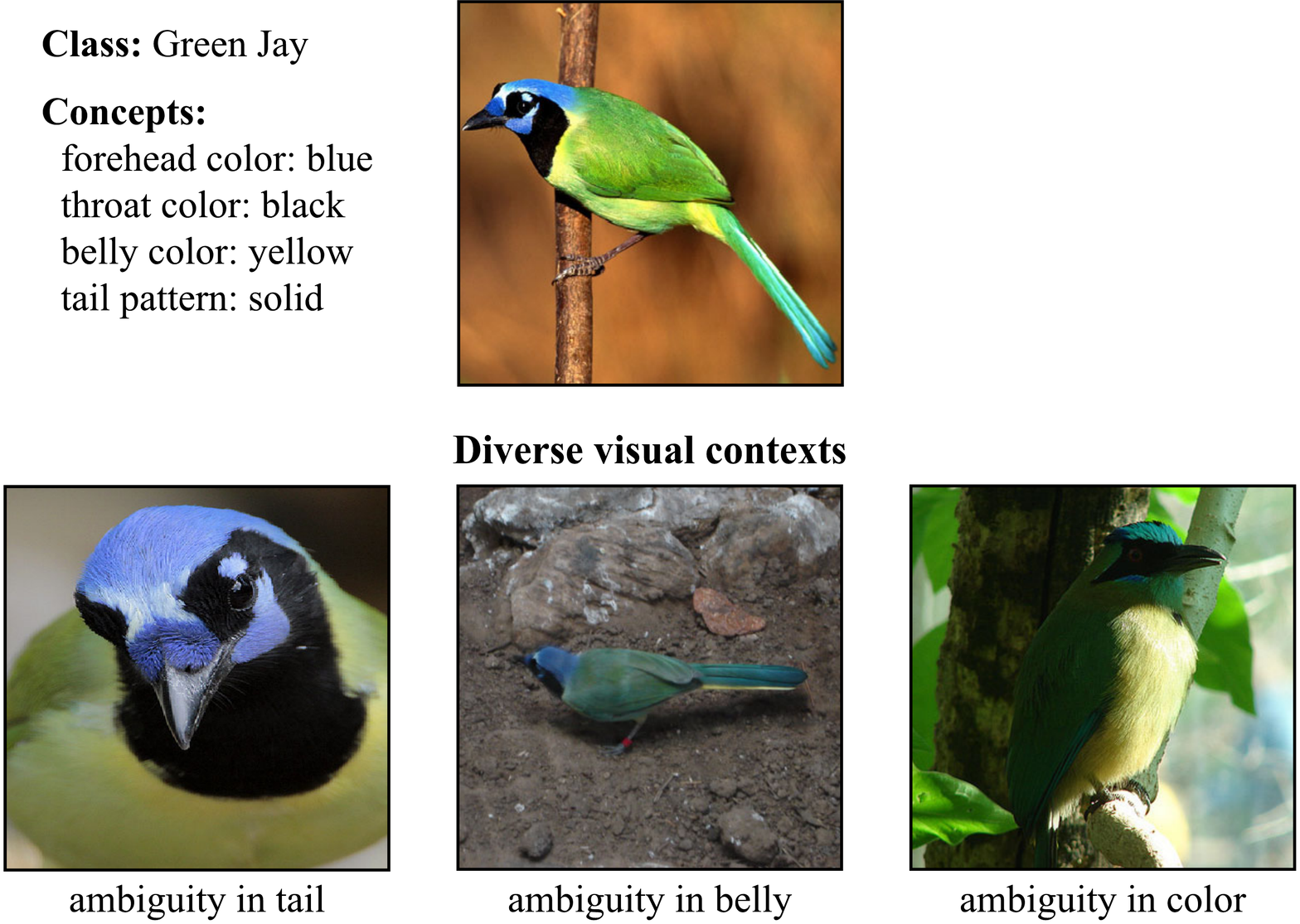}}
\vskip -0.1in
\caption{Examples of ambiguous cases in the existence of the concept. Images have diverse visual contexts, where partial concepts may become invisible and unclear.}
\label{fig:example_uncertain_concept}
\end{center}
\vskip -0.3in
\end{figure}

A concept-based model makes a decision and provides an explanation based on high-level concepts~\cite{koh2020concept, chen2020concept}. Here, the concepts have semantic meanings that align with human understanding and can be expressed via imagery or language~\cite{kim2018interpretability}. Concept Bottleneck Models (CBM)~\cite{koh2020concept}, which are widely used concept-based models, adopt concept prediction in the middle of the decision-making process of the black-box model. In CBM, the final decision is made based on the predicted concepts; thus, the concept prediction serves as an explanation. Accordingly, concept prediction is important for ensuring interpretability in CBM. The concept prediction in CBM is trained as deterministic binary classification by using a dataset that includes concept labels indicating the existence (1) or non-existence (0) of a concept.

However, the existence of a concept is often ambiguous in some cases where the deterministic concept prediction can harm the reliability of the concept explanation.
We conjecture \textit{diverse visual contexts} as the origin of ambiguity in concepts.
Figure~\ref{fig:example_uncertain_concept} illustrates examples belonging to the same class with different visual contexts. In contrast to the top image, the bottom images either lack the tail or belly or exhibit a different color tone, despite belonging to the same class. Deterministic predictions on those properties may harm the faithfulness of the concept explanation. This issue can be further exacerbated when training with discrete concept labels. To alleviate the burden of individually annotating concept labels, images belonging to the same class are normally assigned the same concept labels~\cite{koh2020concept}. Some instances may not actually contain the labeled concepts. Furthermore, data augmentation techniques (\eg random cropping) are commonly used to enhance prediction performance but can introduce diverse visual contexts that lead to the ambiguity issue.

\begin{figure}[t]
\begin{center}
\centerline{\includegraphics[width=\columnwidth]{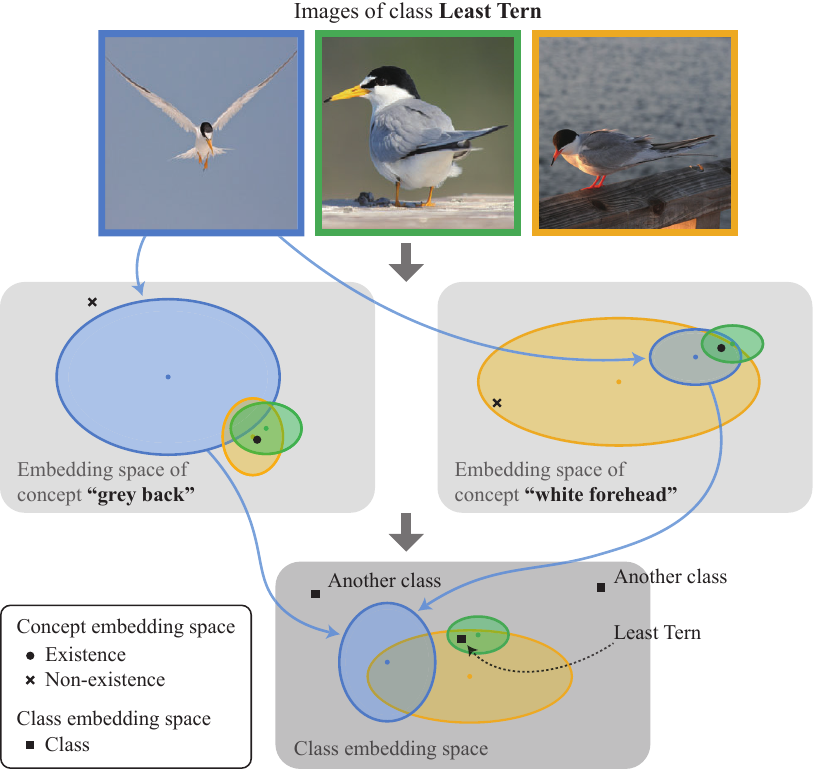}}
\vskip -0.1in
\caption{Probabilistic embeddings in ProbCBM. Images are mapped as probabilistic embeddings in the concept embedding space and the probabilistic concept embeddings are mapped to the class embedding space. Arrows represent those mappings. For simplicity, only arrows of the leftmost image are drawn. The same color represents the same image. The images with more ambiguity in concept prediction are mapped as embeddings with larger ellipses. The anchor points represent the existence and non-existence of the concepts in the concept embedding space and classes in the class embedding space.}
\label{fig:second}
\end{center}
\vskip -0.23in
\end{figure}

To reflect the aforementioned ambiguity in concept prediction, we propose Probabilistic Concept Bottleneck Models (ProbCBM). ProbCBM exploits probabilistic embeddings~\cite{oh2018modeling, shi2019probabilistic, chun2021probabilistic} in the concept embedding space and reflects uncertainty in concept prediction.
Figure~\ref{fig:second} visualizes the concept and class embedding spaces of ProbCBM with examples. Depending on the uncertainty in concept prediction originating from \textit{diverse visual contexts}, ProbCBM maps an image to the concept embeddings with probabilistic distributions, which model concept uncertainties.
The concept embeddings from all concepts are projected to form class embeddings; thus, the final class prediction is derived from concept prediction.

ProbCBM explains its prediction with the predicted concepts and the estimated concept uncertainties, ensuring the reliability of the explanation.
It is also capable of providing class uncertainty drawn by concept uncertainty, which means the class uncertainty can be explained with concept uncertainty.
Through various empirical analyses, we illustrate uncertainty estimation and prediction of ProbCBM. We examine how concept uncertainty varies across different visual contexts and show that the ambiguity introduced by image transformation promotes an increase in uncertainty. We also explore the practical application of estimated uncertainty in human-model interactions in ProbCBM.

Our main contributions are as follows:
\vspace{-5pt}
\begin{itemize}
\item To the best of our knowledge, we pose the ambiguity issue in concept prediction for the first time and address it with uncertainty modeling.
\vspace{-2pt}
\item We propose ProbCBM, an interpretable model that provides explanations based on the concept and its corresponding uncertainty by exploiting probabilistic embeddings.
\vspace{-2pt}
\item We analyze the estimated uncertainty through various experimental results, mainly focusing on the aforementioned origin of ambiguity.

\end{itemize}

\begin{figure*}[t]
\begin{center}
\centerline{\includegraphics[width=0.99\textwidth]{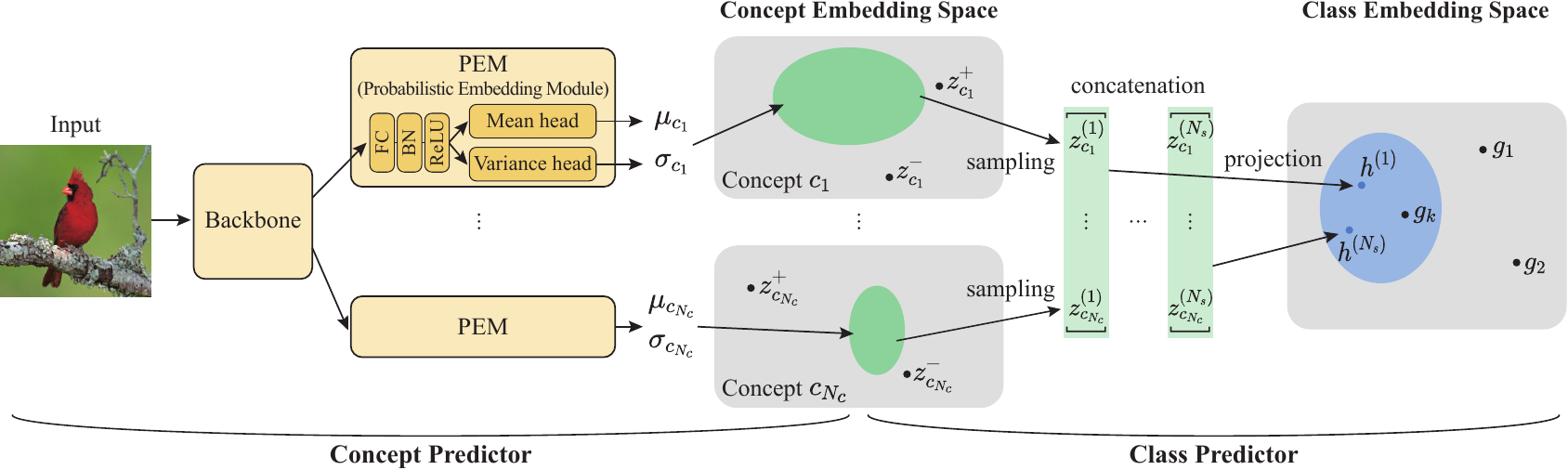}}
\vskip -0.1in
\caption{Prediction flow of ProbCBM. Ellipses represent probabilistic embeddings of a given input.}
\label{fig:flow}
\end{center}
\vskip -0.2in
\end{figure*}

\section{Related Work}\label{sec:relatedwork}
\subsection{Interpretable Neural Networks}
Interpretable neural networks are built to ensure interpretability~\cite{melis2018towards}. One way to build an interpretable prediction process is the use of a concept-based explanation, where models learn human interpretable concepts and make predictions based on the learned concepts~\cite{koh2020concept,chen2019looks}. Since classification models are trained solely with class labels, there have been studies focusing on learning prototypes that represent distinctive properties of each class in an unsupervised manner~\cite{melis2018towards,chen2019looks}.

In contrast, CBM~\cite{koh2020concept} utilizes both concept and class labels. It first predicts the concept for a given input and then proceeds to predict the class. Owing to its simple structure based on human-defined concepts, there have been studies on building improved interpretable models in the framework of CBM~\cite{sarkar2022framework}. To overcome the challenges associated with concept labeling and make use of well-trained models that lack interpretability, Post-hoc CBM (PCBM) was proposed by \citet{yuksekgonul2022post}, allowing the conversion of pre-trained models into CBM. \citet{zarlenga2022concept} posed a trade-off issue between accuracy and interpretability in CBM and proposed a novel model called Concept Embedding Models (CEM) by introducing embeddings to leverage high-level features in task prediction, which represent more information beyond the existence of concepts. Also, some efforts have been made to enhance the reliability of CBM~\cite{havasi2022addressing,marconato2022glancenets}, focusing on an information leakage issue, where unintended information is utilized by the class predictor~\cite{mahinpei2021promises}. However, no research has been performed to mitigate the ambiguity in concept prediction, which is the primary focus of this study.

\subsection{Uncertainty and Probabilistic Embeddings}
Uncertainty modeling is used to improve the interpretability and robustness~\cite{blundell2015weight,pmlr-v48-gal16,kendall2017uncertainties}. Uncertainty is mainly divided into two types: model uncertainty and data uncertainty. Model uncertainty comes from the model parameters, whereas data uncertainty comes from the noise of the data~\cite{pmlr-v48-gal16}.

An approach with probabilistic embeddings is a general method that mainly considers data uncertainty, where the representations of input samples are expressed as probabilistic distributions. \citet{shi2019probabilistic} proposed probabilistic face embeddings to address feature ambiguity in real-world face images. \citet{oh2018modeling} proposed Hedged Instance Embeddings (HIB) and defined the matching probability between a pair via Monte-Carlo estimation. \citet{chun2021probabilistic} extended HIB to the joint embedding space of image and text and solved a cross-modal retrieval problem. We extend HIB to the concept embedding space for concept prediction and build an interpretable model that makes decisions based on the predicted concept embeddings.
\section{Preliminary: Concept Bottleneck Models}\label{sec:preliminary}

CBM~\cite{koh2020concept} consists of two predictors: a concept predictor and a class predictor. Given an input $x$, the concept predictor $g$ maps it to a concept space to predict a set of the existences of concepts $\mathcal{C}$ ($\hat{\mathcal{C}}=g(x)$). The class predictor $f$ estimates the class $y$ from $\hat{\mathcal{C}}$, \ie the concepts predicted by the concept predictor ($\hat{y}=f(\hat{\mathcal{C}})$). CBM's decision-making process is expressed as \mbox{$\hat{y}=f(g(x))$, where} the concept space serves as an interpretable bottleneck for the class prediction. To learn the mapping $g$, the data pairs $(x, \mathcal{C})$ are required. Thus, CBM requires the dataset $\{\left(x^{(i)}, \mathcal{C}^{(i)}, y^{(i)}\right)\}_{i=1}^{N}$ while the conventional classifier that directly maps $x$ to $y$ require the dataset $\{\left(x^{(i)}, y^{(i)}\right)\}_{i=1}^{N}$, where $N$ represents the number of data pairs.

CBM has two strengths with regard to interpretability, which make it useful as an interpretable model. First, it can provide the concept information that it discovers in an input. Because CBM makes the final prediction according to the predicted concept information, the predicted concept information is a valid explanation for the model's decision. Second, concept intervention enables further understanding of the model. In CBM, changes in concept prediction modify the classification result. When the predicted concept is incorrect (not aligned with human understanding), humans can debug the model by intervening in the concept prediction and changing the model's decision. The relationship between a concept and a class can be analyzed by observing the result of concept intervention, which provides counterfactual explanations~\cite{abid2022meaningfully}. Because we build our model, \ie ProbCBM, on the basis of CBM's framework, ProbCBM inherits the strengths of CBM.
\section{Method}\label{sec:method}

\textbf{Overview.}
We propose ProbCBM, an interpretable model that exploits probabilistic embedding in prediction. It provides concept prediction and concept uncertainty as explanations. We extend the probabilistic modeling in HIB~\cite{oh2018modeling} to build ProbCBM. Figure~\ref{fig:flow} shows the overall prediction flow of ProbCBM. Similar to CBM, ProbCBM comprises a concept predictor and a class predictor. The concept predictor generates concept embeddings and the class predictor generates class embeddings from the predicted concept embeddings. With these embeddings, concept and class prediction problems are solved as matching problems with concept and class anchors, respectively. We train ProbCBM with the dataset $\{\left(x^{(i)}, \mathcal{C}^{(i)}, y^{(i)}\right)\}_{i=1}^{N}$.

\subsection{Probabilistic Concept Modeling}
\textbf{Probabilistic concept embedding.}
Given an input $x$, the concept predictor makes probabilistic concept embedding for each concept $c\in\mathcal{C}$, which is formulated as a normal distribution with a mean vector and a diagonal covariance matrix.
\begin{equation}
\label{eq:concept_probabilistic_embedding}  
p(z_c|x)\sim\mathcal{N}(\mu_c, \text{diag}(\sigma_c)),
\end{equation}
where $\mu_c, \sigma_c \in\mathbb{R}^{D_c}$ and $D_c$ represents the dimension of the concept embedding space. $\mu_c$ and $\sigma_c$ are predicted by a probabilistic embedding module (PEM), which is described in Sec.~\ref{sec:method_arch}.
As shown in Figure~\ref{fig:flow}, the generation of probabilistic concept embedding is performed for every concept using a shared backbone and individual PEMs.

\textbf{Concept prediction.}
With $N_s$ representations $\{z_c^{(n)}\}_{n=1}^{N_s}$ sampled from $p(z_c|x)$, the probability of the existence of concept $c$ ($c=1$) is obtained via Monte-Carlo estimation:
\vspace{-5pt}
\begin{equation}
\label{eq:concept_probability_monte_carlo}  
p(c=1|x) \approx \frac{1}{N_s}\sum_{n=1}^{N_s} p(c=1|z_{c}^{(n)}),
\end{equation}
\vspace{-15pt}
\begin{equation}\label{eq:prob_concept}
\begin{aligned}
    p(c&=1|z_{c}^{(n)})\\
    &=s\left(a\left(||z_{c}^{(n)}-z_{c}^-||_2-||z_{c}^{(n)}-z_{c}^+||_2\right)\right),
\end{aligned}
\end{equation}
where $a>0$ is a learnable parameter and $s(\cdot)$ represents a sigmoid function. We define trainable anchor points $z_{c}^+$ and $z_{c}^-$ in $\mathbb{R}^{D_c}$, which represent the existence and non-existence of concept $c$, respectively. For a given sampled representation $z_{c}^{(n)}$, the probability that the concept exists is defined by the Euclidean distance with $z_{c}^+$ and $z_{c}^-$. If $z_{c}^{(n)}$ becomes closer to $z_{c}^+$ than $z_{c}^-$, the probability of existence increases, and if it becomes farther away, the probability decreases. More explanations on design of concept prediction are provided in Appendix~\ref{sec:appendix_anchors}.

\subsection{Probabilistic Class Modeling}
\textbf{Class embedding.}
The class predictor generates a class embedding from the concept embeddings. We concatenate sampled concept representations from all concepts $\mathcal{C}=\{c_1, c_2, \dots, c_{N_c}\}$, where $N_c$ represents the number of concepts. The concatenation is projected to the class embedding space with a fully connected (FC) layer.
\begin{equation}\label{eq:proj_class}
    h^{(n)}=\mathbf{w}^T\left(\left[z_{c_1}^{(n)}, z_{c_2}^{(n)}, \dots, z_{c_{N_c}}^{(n)}\right]\right) + \mathbf{b},
\end{equation}
where $h^{(n)}$ is a class representation and $\mathbf{w}$ and $\mathbf{b}$ represent the weight and bias of the FC layer, respectively.

\textbf{Class prediction.}
The logit for class $k$ is defined by the Euclidean distance between the class embedding for the image and a trainable anchor point for class $k$, $g_k \in \mathbb{R}^{D_y}$ where $D_y$ represents the dimension of the class embedding space. We obtain the class probabilities by applying softmax to the logits for overall classes. The classification probability is obtained via Monte-Carlo estimation:
\begin{equation}\label{eq:prob_class_mc}
    p(y_k=1|x)\approx \frac{1}{N_s}\sum_n^{N_s}\frac{\exp(-d\| h^{(n)}-g_k\|_2)}{\sum_{k'}{\exp(-d\| h^{(n)}-g_{k'}\|_2)}},
\end{equation}
where $d>0$ is a learnable parameter.

\subsection{Training and Inference}
\textbf{Training objective.}
We use a binary cross-entropy loss ($\mathcal{L}_\text{BCE}$) with the concept probability (Eq.~\ref{eq:concept_probability_monte_carlo}).
Following HIB~\cite{oh2018modeling}, we additionally use a KL divergence loss between the predicted concept embedding distributions and the standard normal distribution. This prevents the variances from collapsing to zero and makes the distribution $\mathcal{N}(\mu_c, \text{diag}(\sigma_c))$ have only salient information for predicting the probability that the concept $c$ exists.
\begin{equation}\label{eq:loss_kl}
    \mathcal{L}_\text{KL}(c)=\text{KL}\left(\mathcal{N}(\mu_c, \text{diag}(\sigma_c))||\mathcal{N}(0, I)\right).
\end{equation}
Thus, the overall training loss for the concept predictor is expressed as:
\begin{equation}\label{eq:loss_total}
    \mathcal{L}_\text{concept} = \mathcal{L}_\text{BCE} + \lambda_\text{KL}\mathcal{L}_\text{KL},
\end{equation}
where $\lambda_\text{KL}$ is a balancing factor.

We use a cross-entropy loss for training class predictor ($\mathcal{L}_\text{class}$).

\textbf{Training scheme.}\label{sec:method_training}
We train the concept predictor and class predictor separately. First, the concept predictor is trained with $\mathcal{L}_\text{concept}$. Then, the class predictor is trained with $\mathcal{L}_\text{class}$ using the concept embeddings predicted by the concept predictor. During the training of the class predictor, a sampled concept embedding $z_{c}^{(n)}$ is replaced with the concept anchor of a ground-truth concept label ($z_{c}^+$ and $z_{c}^-$ for positive and negative labels, respectively) with the predefined probability $p_\text{replace}$. This prevents the class predictor from learning with incorrect concepts, improving the final classification performance and the reliability of the class predictor. See Algorithm~\ref{alg:training_scheme} in the Appendix for details.

\textbf{Inference.}
Inference can be done by approximating the probabilities via Monte-Carlo sampling (Eqs.~\ref{eq:prob_concept} and \ref{eq:prob_class_mc}) or using $\mu_c$ as $z_c$ without sampling. The inference method is discussed in Sec~\ref{sec:exp_cls_result}.

\subsection{Derivation of Uncertainty}
\vspace{-2pt}
ProbCBM leverages probabilistic modeling, enabling us to estimate uncertainty directly from the predicted probabilistic distribution without the need for sampling. We quantify uncertainty using the determinant of the covariance matrix, which represents the volume of the probabilistic distribution. Because the distribution of the concept embedding is parameterized with a diagonal covariance matrix, the uncertainty of each concept $c$ can be calculated as the geometric mean of the diagonal elements $\sigma_{c}$. The class embedding is a linear transformation of the concatenation of concept embeddings, and the class embeddings follow $\mathcal{N}(\mathbf{w}^T\mu+\mathbf{b}, \mathbf{w}^T\Sigma \mathbf{w})$, where $\mu=\left[\mu_{c_1}, \mu_{c_2}, ..., \mu_{c_{N_c}}\right]$ and $\Sigma=\text{diag}\left(\left[\sigma_{c_1}, \sigma_{c_2}, ..., \sigma_{c_{N_c}}\right]\right)$. Hence, the determinant of $\mathbf{w}^T\Sigma \mathbf{w}$ serves as an uncertainty measure of class prediction.

\vspace{-2pt}
\subsection{Architecture}\label{sec:method_arch}
\vspace{-2pt}
As shown in Figure~\ref{fig:flow}, with a shared backbone (\eg ResNet~\cite{he2016deep}), there is one PEM for each concept. PEM predicts the mean and diagonal covariance vector for the corresponding concept from a feature map extracted from the backbone. Following the work of \citet{chun2021probabilistic}, we use self-attention-based mean and variance head submodules in PEM. To reduce the feature dimension, we add an FC layer followed by batch normalization~\cite{ioffe2015batch} and ReLU activation~\cite{nair2010rectified} before the mean and variance head modules.

\begin{figure}[t]
\begin{center}
\centerline{\includegraphics[width=0.98\columnwidth]
{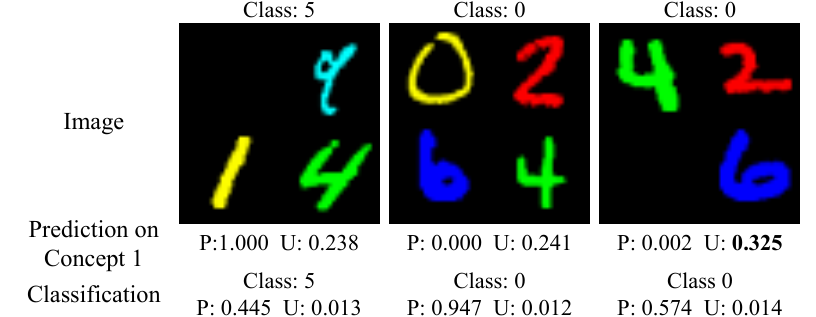}}
\vskip -0.12in
\caption{Examples of ProbCBM's prediction in diverse visual contexts. P and U represent the probability and uncertainty, respectively. The highest value of concept uncertainty is bolded.}
\label{fig:example_mnist1}
\end{center}
\vskip -0.27in
\end{figure}
\begin{figure}[t]
\begin{center}
\centerline{\includegraphics[width=\columnwidth]
{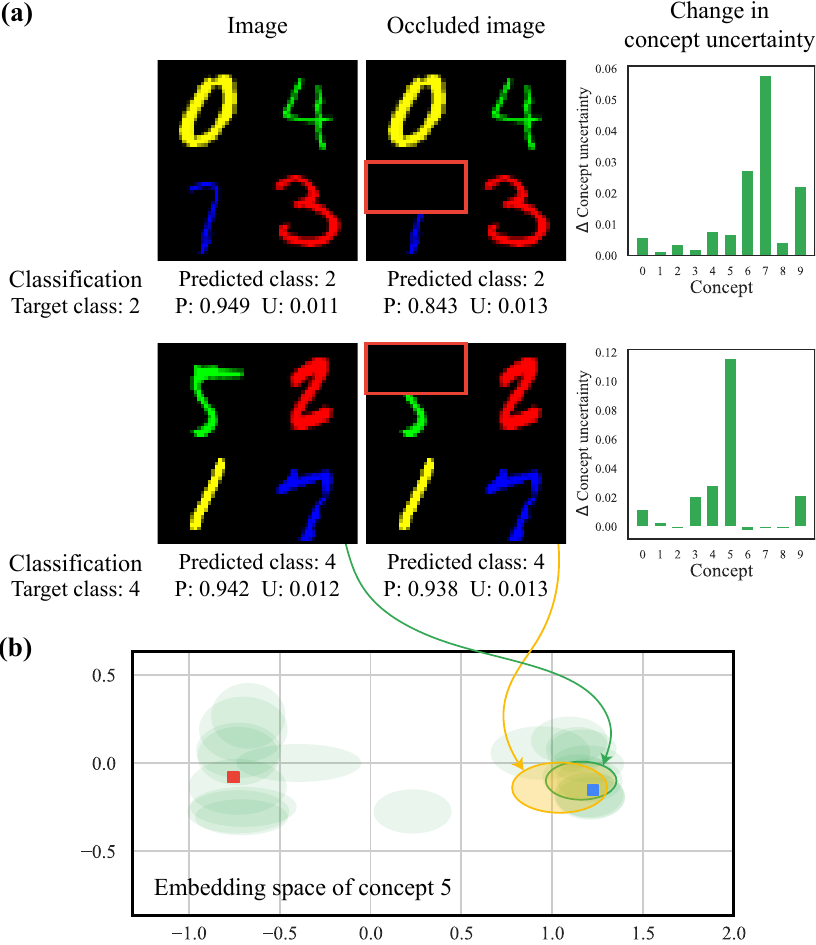}}
\vskip -0.05in
\caption{(a) Examples of changes in the concept and class uncertainties after occlusion of parts of images. Red bounding boxes denote the occluded parts. P and U represent the probability and uncertainty, respectively. (b) Visualization of the embedding space of concept 5. Blue and red dots present positive and negative anchors, respectively. The probabilistic concept embeddings are visualized as green and yellow ellipses. Blue and red points represent positive and negative anchor points, respectively.}
\label{fig:example_mnist}
\end{center}
\vskip -0.1in
\end{figure}

\section{Experiments}\label{sec:exp}
\vspace{-2pt}
We evaluate ProbCBM with one synthetic dataset and two real-world datasets. With these datasets, we demonstrate how ProbCBM effectively models uncertainty under diverse visual contexts presented in raw images. Additionally, we conduct an analysis to examine the ambiguity induced by image transformations.

\vspace{-2pt}
\subsection{Analysis with Synthetic Dataset}
\vspace{-2pt}
\subsubsection{Dataset and experimental setting} \label{sec:exp_dataset_mnist}
\vspace{-2pt}
\textbf{Dataset.}
We create synthetic data using the MNIST dataset \cite{lecun2010mnist} which contains images of 10 digits. We utilize each digit as a single concept. We first divide the 10 digits into five groups: (0, 1), (2, 3), (4, 5), (6, 7), and (8, 9). We then generate a single image with four digits from four different groups and annotate a new class label out of 12 classes, depending on the combination of digits. The digits in the same group are colorized the same color. To add diversity to the concept combinations for each class, we randomly drop one of the four digits in the images. There is no image that can be considered as multiple classes owing to the drop because no class shares more than three concepts with any other class. To simulate ambiguity in the training data, we use cutout~\cite{devries2017improved} augmentation method. The synthetic dataset provides instance-level concept annotations. See Appendix~\ref{sec:appendix_dataset} for details.

\textbf{Setting.}
We use a shallow convolutional network consisting of two convolutional layers with $3 \times 3$ filters, followed by batch normalization and ReLU activation as a backbone. ProbCBM achieves accuracies of 99.8\% and 99.8\% for concept prediction and classification, respectively. See Appendix~\ref{sec:appendix_exp_detail} for experimental details.

\begin{table}[t]
\setlength{\tabcolsep}{4.3pt}
\caption{Prediction accuracy for the CUB and AwA2 datasets. Results include mean values with standard deviation from three experiment repetitions. ``w/o prob." denotes ProbCBM with deterministic embeddings. ``w/o sampling" denotes inference without sampling.}
\begin{center}
\begin{small}
\begin{sc}
\begin{tabular}{clcc}\toprule
Data&Method&Concept &Class \\\midrule
\multirow{10}{*}{CUB} & \multicolumn{2}{l}{~~Image: $224 \times 224$}\\
&CBM &0.950$\pm$0.001 &0.670$\pm$0.006 \\
&ProbCBM &0.949$\pm$0.001 &0.680$\pm$0.004 \\
&~~w/o prob. &0.950$\pm$0.001 &0.677$\pm$0.004 \\
&~~w/o sampling &0.949$\pm$0.001 &0.679$\pm$0.003 \\
\cmidrule{2-4}
& \multicolumn{2}{l}{~~Image: $299 \times 299$}\\
&CBM & 0.956$\pm$0.001 & 0.708$\pm$0.006\\
&CEM & 0.954$\pm$0.001 & 0.759$\pm$0.002\\
&ProbCBM & 0.956$\pm$0.001& 0.718$\pm$0.005\\
\midrule
\multirow{4}{*}{AwA2} &CBM &0.975$\pm$0.001 &0.877$\pm$0.004 \\
&ProbCBM &0.975$\pm$0.000 &0.880$\pm$0.002 \\
&~~w/o prob. &0.975$\pm$0.000 &0.880$\pm$0.002 \\
&~~w/o sampling &0.975$\pm$0.000 &0.880$\pm$0.001 \\
\bottomrule
\end{tabular}\label{tab:perf_cub}
\end{sc}
\end{small}
\end{center}
\vskip -0.2in
\end{table}

\subsubsection{Understanding estimated uncertainty}
\vspace{-1pt}
\textbf{Ambiguity in diverse visual contexts.}
We compare the estimated uncertainties of multiple images which have diverse visual contexts. Figure~\ref{fig:example_mnist1} shows examples of ProbCBM's predictions for concept 1, where the left two images are predicted with smaller uncertainties, and the rightmost image is predicted with a larger uncertainty. Because the leftmost image contains concept 1, ProbCBM predicts that concept 1 exists, with a small uncertainty. Additionally, ProbCBM confidently predicts that concept 1 does not exist in the center image. Note that concepts 0 and 1 are in the same group; thus, the existence of concept 0 implies the non-existence of concept 1. The rightmost image contains neither concept 0 nor concept 1; thus, the estimated uncertainty for concept 1 is large. This indicates that the estimated concept uncertainty depends on the visual context. The estimated concept uncertainty is large when there is ambiguity in the prediction of the corresponding concept, leading to large uncertainty in classification.

\textbf{Introducing ambiguity via transformation.}
We artificially introduce ambiguity in the data by occluding the upper half region of one of four digits in images. Figure~\ref{fig:example_mnist}(a) shows examples of the changes in concept uncertainty after the occlusion of part of the images. When part of a digit is occluded, the uncertainty of the corresponding concept increases significantly (concept 7 in $1^{st}$ row and concept 5 in $2^{nd}$ row), increasing the class uncertainty.

\textbf{Visualization in concept embedding space.}
We visualize the predicted probabilistic concept embeddings and positive and negative anchors of concept 5, as shown in Figure~\ref{fig:example_mnist}(b). We use principal component analysis (PCA)~\cite{pearson1901liii} to visualize them in a two-dimensional (2D) space. Each ellipse represents the predicted concept embedding of an image, whose size denotes the uncertainty. The embeddings of the samples are distributed near the positive or negative anchors and have various uncertainty values. The concept embedding of the occluded image is represented by an ellipse larger than that of the original image. This indicates that ProbCBM successfully models concept uncertainty by using probabilistic embeddings.

\begin{figure}[t]
\begin{center}
\centerline{\includegraphics[width=0.98\columnwidth]{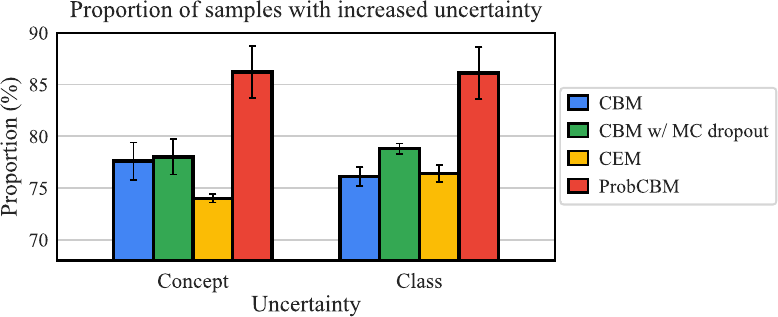}}
\vskip -0.15in
\caption{Proportion (\%) of samples whose uncertainty increases after occlusion for the CUB dataset. The experiments are conducted with an image size of $299 \times 299$. Results include mean values with standard deviation from three experiment repetitions.}
\label{fig:uncertainty_metric}
\end{center}
\vskip -0.32in
\end{figure}
\begin{figure*}[t]
\begin{center}
\centerline{\includegraphics[width=\textwidth]{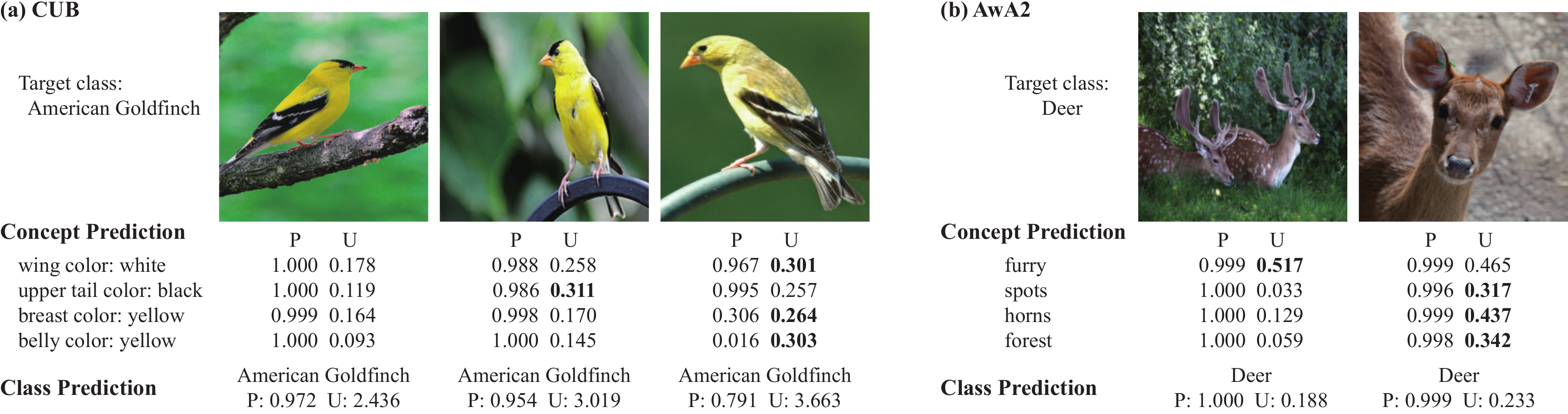}}
\vskip -0.14in
\caption{Examples of ProbCBM's prediction in diverse visual contexts for the (a) CUB and (b) AwA2 datasets. P and U represent the probability and uncertainty, respectively. The highest value of uncertainty in each concept is bolded.}
\label{fig:exp_example}
\end{center}
\vskip -0.22in
\end{figure*}

\vspace{-1pt}
\subsection{Analysis with Real-world Datasets}
\vspace{-1pt}

\subsubsection{Datasets and experimental setting}
\vspace{-1pt}
\textbf{CUB.}
Caltech-UCSD Birds-200-2011 (CUB) dataset \cite{wah2011caltech} contains 11,788 images from 200 bird species, which are annotated with 312 attributes. \citet{koh2020concept} denoised the attribute annotations and 112 attributes remained, where the images belonging to the same class have the same attribute annotations. We use the attribute annotations and dataset splits that \citet{koh2020concept} provided. The attribute annotations are used as concepts.

\textbf{AwA2.}
Animal with Attributes (AwA2) dataset~\cite{xian2018zero} contains 37,322 images from 50 animal classes, which are annotated with 85 attributes. Because the list of attributes includes invisible attributes (\eg fast, slow), we retain 45 attributes that are visible in images and use them as concepts. See Appendix~\ref{sec:appendix_dataset} for the remaining attribute labels.

\textbf{Setting.}
We use ResNet18~\cite{he2016deep} as a backbone, which is pretrained with ImageNet-1K~\cite{russakovsky2015imagenet}. We use color jittering, random horizontal flips, and random resized cropping for data augmentation. Note that we apply the same data augmentation to all methods, which is a weaker level of augmentation than that used in other works (CBM and CEM~\cite{zarlenga2022concept}), to reduce the distortion of concept labels due to data augmentation\footnote{This makes a difference in the reported performance of prediction of other methods.}. For most experiments, the images are resized to $224 \times 224$. We additionally conduct experiments with the image size of $299 \times 299$ to compare ProbCBM with other methods for the CUB dataset (Sec.~\ref{sec:exp_cls_result}). See Appendix~\ref{sec:appendix_exp_detail} for more details.

\subsubsection{Comparisons}\label{sec:exp_cls_result}
\vspace{-1pt}
\textbf{Prediction performance.}
It is important to enhance the reliability of the model while avoiding a drop in prediction performance. To evaluate this aspect, we compare the concept and class prediction performances of ProbCBM with CBM and CEM. CEM is chosen as a comparison method because it also utilizes concept embeddings.

The results are presented in Table~\ref{tab:perf_cub}. ProbCBM demonstrates superior classification accuracy compared to CBM while maintaining a comparable concept prediction accuracy. This indicates that introducing uncertainty with probabilistic embedding does not drop prediction performance. ProbCBM achieves lower class prediction performance but slightly higher concept prediction performance compared to CEM. This is due to their distinct training approaches. In CEM, concept embeddings are trained to represent other information in addition to the existence or non-existence of concepts, aiming to improve classification performance through joint training of concept and class predictors. In contrast, ProbCBM focuses on training concept embeddings to represent uncertainties without explicit consideration of classification through sequential training. They prioritize different aspects of prediction.  Additionally, it is worth noting that CEM has a larger number of parameters (15.6M) compared to ProbCBM (12.5M). The comparison of prediction performances with other models shows that ProbCBM does not compromise prediction performance as well as provides uncertainty-based interpretation.

To examine the effect of probabilistic embedding, we also evaluate the performance of the model that exploits deterministic concept and class embeddings, which we refer to as ``w/o prob.," as shown in Table~\ref{tab:perf_cub}. The classification accuracy of ProbCBM is higher than or comparable to that of the model exploiting deterministic embeddings.

\textbf{Efficient inference.}
Inference of ProbCBM can be done with and without Monte-Carlo sampling. To reduce the computational cost for sampling, we can use the predicted means instead of sampled representations. As shown in Table~\ref{tab:perf_cub}, the difference between inference with sampling and inference without sampling is minor. The anchor points can be considered as the distributions following a normal distribution with zero variance. Thus, using the means can be viewed as directly using the distance between two distributions \textendash~probabilistic class embeddings and class anchors \textendash~in prediction.

\textbf{Uncertainty estimation.}
To evaluate the capability of estimating uncertainty, we intentionally design scenarios that would induce an increase in uncertainty and quantitatively compare the ability to detect this increase. Motivated by a common evaluation protocol~\citep{oh2018modeling,shi2019probabilistic}, we occlude a patch of size $64 \times 64$ from the center of images. This occlusion induces a loss of information, resulting in an increase in uncertainty. We then analyze the proportion of samples that exhibit higher uncertainty than before. In addition to CBM and CEM, we compare our method with MC dropout~\citep{pmlr-v48-gal16}, a well-known approach for uncertainty estimation, by adopting it to CBM (CBM with MC dropout). Uncertainties in the models, except for ProbCBM, are estimated using the entropy of the predicted concept and class probabilities.

Figure~\ref{fig:uncertainty_metric} depicts the proportion of samples exhibiting increased uncertainty after occlusion. The results clearly demonstrate that ProbCBM is effective in detecting an increase in both concept and class uncertainties across a wide range of samples. While MC dropout improves the detection of increased uncertainty in concept and class, its performance is inferior to that of ProbCBM. Notably, CEM, despite employing concept embeddings, diverges significantly in uncertainty estimation due to disparate objectives and strategies compared to ProbCBM. These findings affirm the effectiveness of ProbCBM in uncertainty estimation.

\begin{figure}[t]
\begin{center}
\centerline{\includegraphics[width=\columnwidth]
{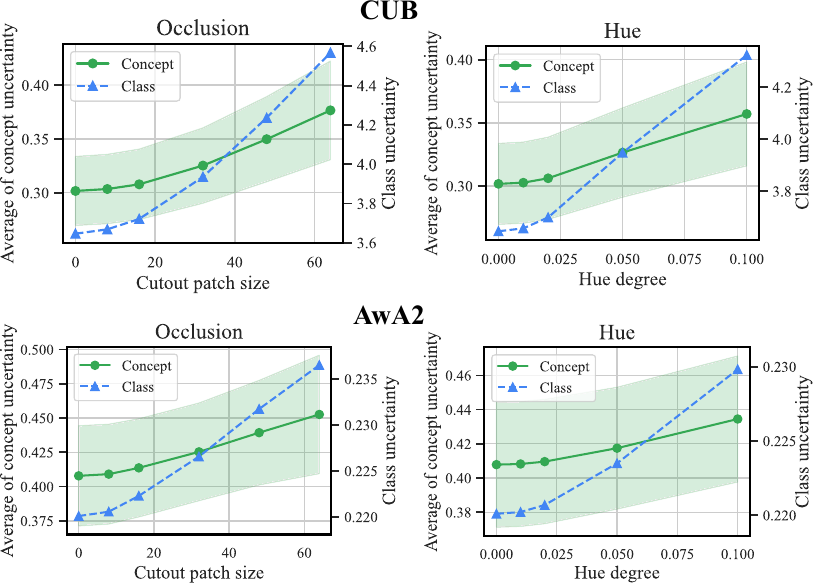}}
\vskip -0.15in
\caption{Changes in uncertainty after the introduction of ambiguity via image transformations to various degrees.
For the concept uncertainty, line and shade represent the mean and standard deviation over all concepts, respectively.}
\label{fig:plot_uncertainty_cutout}
\end{center}
\vskip -0.25in
\end{figure}
\subsubsection{Understanding estimated uncertainty}\label{sec:exp_real_uncertainty}
\textbf{Ambiguity in diverse visual contexts.}
Figure~\ref{fig:exp_example} shows examples of images belonging to the same class in diverse visual contexts. Regarding Figure~\ref{fig:exp_example}(a), the concepts are well visible in the leftmost image; thus, the estimated concept uncertainties are small, inducing small uncertainty in class prediction. The center and rightmost images mainly show the bird's belly and back, respectively, which exhibit different uncertainties.
Regarding Figure~\ref{fig:exp_example}(b), the left image contains spots and horns. The right image does not contain them, but the concept predictor is trained to predict images with deer to have horns and spots; thus, the predictions for these concepts exhibit high uncertainties. Note that class-level concept annotations are used to train ProbCBM, where images belonging to the same class are annotated with the same concepts.

\textbf{Introducing ambiguity via transformation.}
We artificially add ambiguity to images by applying transformations: cutout and hue. We cut out a square patch from an image or change the hue of an image. 
Because the sets of defined concepts in the CUB and AwA2 datasets contain descriptions of parts of objects (\eg neck, leg) and colors (\eg wing color), these transformations can remove or distort information about concepts in the images. As shown in Figure~\ref{fig:plot_uncertainty_cutout}, a larger size of the patch cut out or degree of change in the hue makes the bigger correspond to larger increases in the concept and class uncertainties, indicating that uncertainty increases as the ambiguity increases.

\begin{figure}[t]
\begin{center}
\centerline{\includegraphics[width=\columnwidth]
{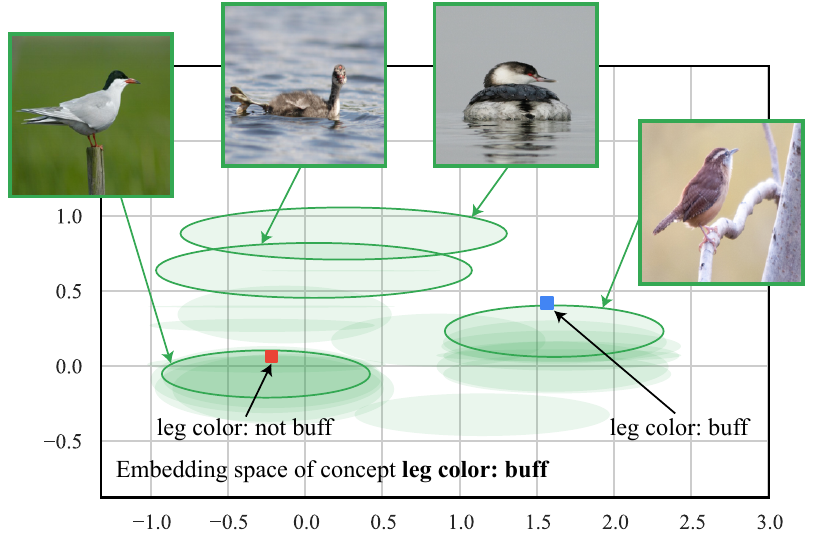}}
\vskip -0.1in
\caption{Visualization of the embedding space of the concept \texttt{leg color: buff}. Each green ellipse represents the concept embedding of the corresponding sample.}
\label{fig:concept_embedding}
\end{center}
\vskip -0.2in
\end{figure}

\textbf{Visualization in embedding space.}
We visualize predicted probabilistic embeddings in the 2D space via PCA. Figure~\ref{fig:concept_embedding} shows the embedding space of the concept \texttt{leg color: buff}. The image with the buff-colored leg is located near the positive anchor, and the image with the red-colored leg is located near the negative anchor. Two images where the leg is invisible are visualized as larger ellipses between two anchors, representing larger uncertainties. Such a large concept uncertainty can induce a large uncertainty in classification, as shown in Figure~\ref{fig:appendix_embedding} in the Appendix.

\textbf{Correlation between uncertainty and performance.}
\begin{figure}[t]
\begin{center}
\centerline{\includegraphics[width=\columnwidth]{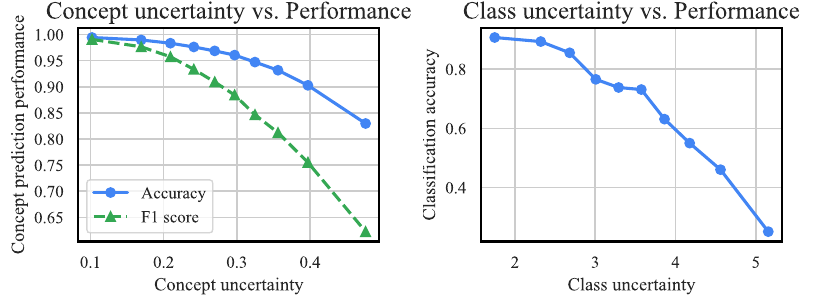}}
\vskip -0.1in
\caption{Correlation between performance and uncertainty for the CUB dataset. A point represents one group of samples with similar uncertainties and the x-axis indicates the median uncertainty for each group.
}
\label{fig:perf_uncertainty}
\end{center}
\vskip -0.2in
\end{figure}
The estimated uncertainty can be used to estimate the probability of failure in prediction. We divide the test data into 10 groups according to the degree of uncertainty, where each group has the same number of samples, and evaluate the prediction performance for each group. As shown in Figure~\ref{fig:perf_uncertainty}, groups with larger uncertainties exhibit worse performances for concept and class prediction.

\subsubsection{Concept intervention}\label{sec:concept_intervention}
\begin{figure}[t]
\begin{center}
\centerline{\includegraphics[width=0.98\columnwidth]
{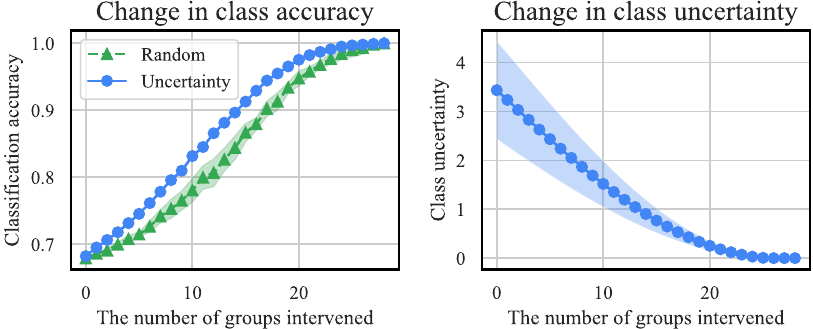}}
\vskip -0.11in
\caption{Results of concept intervention for the CUB dataset. Intervention in a random order is conducted five times and is visualized with the means and standard deviations. On the right plots, intervention is conducted in an uncertainty-based order, and the means and standard deviations for all test samples are shown.}
\label{fig:concept_intervention}
\end{center}
\vskip -0.28in
\end{figure}

Concept intervention is performed to revise incorrect concept predictions for modifying model prediction~\cite{koh2020concept}. In ProbCBM, concept intervention can be done by replacing sampled concept points with anchors of the ground-truth concept. Following \citet{koh2020concept}, we group the concepts belonging to a similar category and intervene in the concepts in the same group at once (28 groups for the CUB dataset). As shown in Figure~\ref{fig:concept_intervention}, intervening incorrectly predicted concepts, even in random order, consistently increases the classification accuracy, which eventually reaches 100\%. This indicates that, during debugging, humans can fix wrong model predictions by intervening in incorrect concepts in ProbCBM.

\begin{figure}[t]
\begin{center}
\centerline{\includegraphics[width=\columnwidth]
{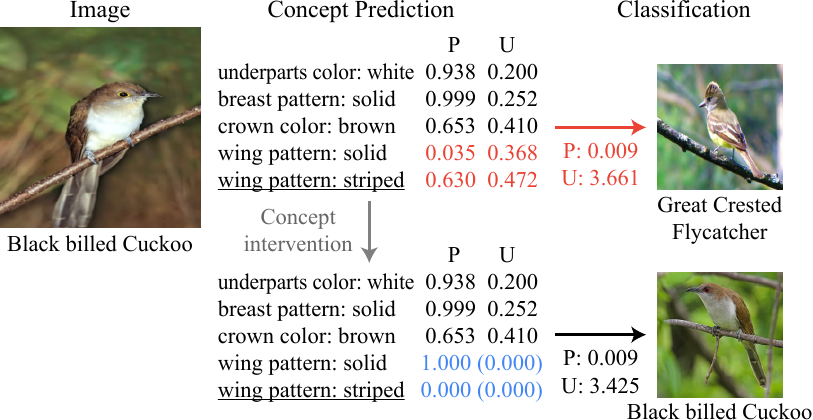}}
\vskip -0.12in
\caption{Examples of concept intervention in ProbCBM. P and U represent the probability and uncertainty, respectively. The concept labels for most concepts are positive, except those that are underlined. Red text denotes wrong predictions and blue text denotes the change after concept intervention.}
\label{fig:example_cub}
\end{center}
\vskip -0.25in
\end{figure}

Practically, it is challenging to ascertain the correctness of concept prediction for all concepts. Therefore, it is necessary to develop an efficient strategy that can effectively identify concepts to intervene first to quickly revise the final predictions. Estimated uncertainty serves as a valuable tool for efficient concept intervention~\cite{shin2022a,chauhan2022interactive,sheth2022learning}. The concepts are intervened in descending order of the maximum uncertainty values of concepts within a group. Figure~\ref{fig:concept_intervention} demonstrates that the order based on uncertainty yields a faster improvement in class accuracy than random orders in ProbCBM. Figure~\ref{fig:example_cub} provides an example of concept intervention where we intervene in the concepts related to the wing pattern, which exhibit the highest uncertainties. Consequently, the class prediction is fixed and the class uncertainty decreases.

Likewise, we can make more certain class predictions via concept intervention in ProbCBM. Intervening in concepts makes concept embeddings deterministic; thus, the uncertainty of the intervened concepts becomes zero. Concept intervention in ProbCBM includes not only intervening in the prediction of the existence of concepts but also intervening in the concept uncertainties, affecting the class uncertainty. As shown in Figure~\ref{fig:concept_intervention}, the class uncertainty decreases with intervention in more groups of concepts.

\vspace{-1pt}
\section{Limitations}\label{sec:limitation}
ProbCBM is built upon CBM, aiming to enhance the reliability of concept prediction by incorporating uncertainty. Consequently, it inherits some limitations from CBM. One such limitation is the necessity of concept labels. Thus, in CBM, the accuracy of concept prediction is heavily influenced by the quality of human-annotated concept labels. ProbCBM utilizes concept labels, but by introducing uncertainty to address the ambiguity issue, we can reduce the dependency on the correctness of concept labels. Another limitation is information leakage~\cite{mahinpei2021promises}, the usage of unintended information in task prediction. The motivation for the concept bottleneck is to build an interpretable prediction process using task prediction only based on human-interpretable concepts. However, the leakage of information that is not interpretable by humans can harm reliability. While we improved reliability in terms of uncertainty, information leakage is also a problem that needs to be resolved to enhance reliability in the framework of CBM.

\section{Conclusion}\label{sec:conclusion}
In this paper, we first pose the ambiguity issue that can harm the reliability of concept prediction in the concept bottleneck model. We propose ProbCBM which can successfully reflect the ambiguity as concept uncertainty by introducing probabilistic embeddings. Since the class embedding is derived from the probabilistic concept embedding, ProbCBM is also able to provide uncertainty in class prediction. Various analysis has been presented to help understand ProbCBM and show the improved reliability of the explanations.

\section*{Acknowledgements}

The authors would like to express their appreciation to Sanghyuk Chun and Sangdoo Yun at NAVER AI Lab for their significant help in the ideation and experimental designs of this work. Many aspects of this work were devised and carried out during the first author's internship at Naver AI Lab, utilizing the Naver Smart Machine Learning (NSML) platform~\cite{kim2018nsml}.

This work was supported by the National Research Foundation of Korea (NRF) grants funded by the Korea government (Ministry of Science and ICT, MSIT) (2022R1A3B1077720 and 2022R1A5A708390811), Institute of Information \& Communications Technology Planning \& Evaluation (IITP) grants funded by the Korea government (MSIT) (2022-0-00959 and 2021-0-01343: AI Graduate School Program, SNU), and the BK21 FOUR program of the Education and Research Program for Future ICT Pioneers, Seoul National University in 2023.

\bibliography{egbib}
\bibliographystyle{icml2023}

%%%%%%%%%%%%%%%%%%%%%%%%%%%%%%%%%%%%%%%%%%%%%%%%%%%%%%%%%%%%%%%%%%%%%%%%%%%%%%%
%%%%%%%%%%%%%%%%%%%%%%%%%%%%%%%%%%%%%%%%%%%%%%%%%%%%%%%%%%%%%%%%%%%%%%%%%%%%%%%
% APPENDIX
%%%%%%%%%%%%%%%%%%%%%%%%%%%%%%%%%%%%%%%%%%%%%%%%%%%%%%%%%%%%%%%%%%%%%%%%%%%%%%%
%%%%%%%%%%%%%%%%%%%%%%%%%%%%%%%%%%%%%%%%%%%%%%%%%%%%%%%%%%%%%%%%%%%%%%%%%%%%%%%
\newpage
\appendix
\onecolumn

\section{Usage of two anchors for concept prediction}\label{sec:appendix_anchors}

Concept prediction can be exclusively performed using a positive anchor, where the distance between the predicted concept embedding and the positive anchor represents the probability of the concept's non-existence. This approach trains the model to generate embeddings close to the positive anchor for samples with the concept (positive samples) and far from the positive anchor for samples without the concept (negative samples).  As a result, the magnitude of $\sigma$ of negative samples significantly exceeds that of positive samples. Consequently, the volume of embedding becomes an unreliable estimate of uncertainty when the positive anchor is exclusively used. On the other hand, using separate positive and negative anchors provides targets for both proximity and distance, allowing $\sigma$ of probabilistic embeddings to represent uncertainty reliably.

\section{Discussion on data augmentation and ambiguity}\label{sec:appendix_augmenatation}

CBM heavily relies on concept labels, and the accuracy and reliability of concept prediction are highly contingent on the quality of human-annotated concept labels. In addition to the incompleteness inherent in concept labels, data augmentation can exacerbate this incompleteness. Concept labels are more susceptible to being influenced by data augmentation techniques compared to class labels. For instance, in datasets like CUB, there are concepts related to specific object parts, and random cropping can inadvertently remove or obscure those parts. Similarly, concepts related to colors can be affected by strong color jittering, resulting in distorted color representation. Therefore, it is vital to employ meticulous data augmentation techniques that preserve the integrity of concept labels, considering different types of concepts.

Likewise, data augmentation introduces a wide range of images with diverse visual contexts, which can result in ambiguity issues. The introduction of ambiguity brings a challenge to reliable concept prediction, as we posed in this paper. ProbCBM effectively tackles the ambiguity issue by modeling and estimating uncertainty. By doing so, it mitigates the adverse effects of diverse visual contexts, thereby enhancing the reliability of concept prediction. Through the incorporation of uncertainty estimation, ProbCBM offers a dependable approach to concept prediction in the presence of ambiguity.

\section{Details of datasets}\label{sec:appendix_dataset}
\subsection{Synthetic dataset based on MNIST dataset}
\begin{figure*}[h]
\begin{center}
\centerline{\includegraphics[width=0.8\textwidth]{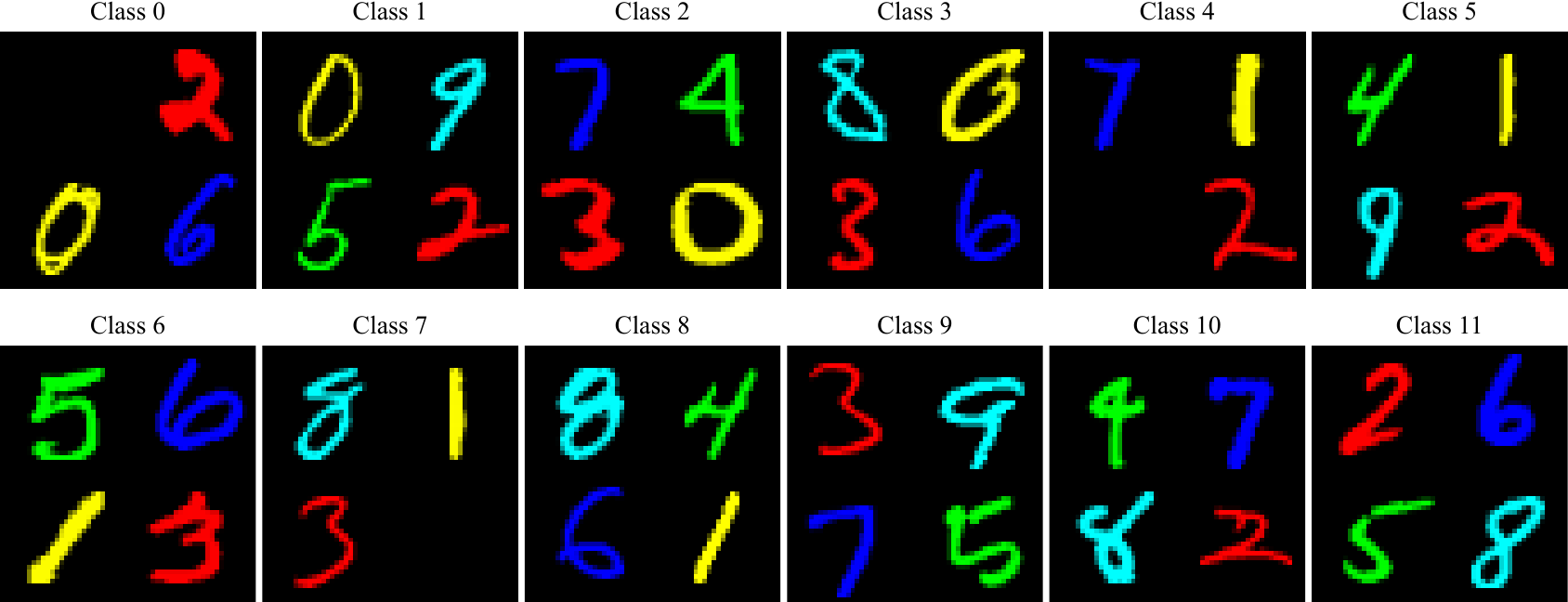}}
\caption{Example images from the synthetic dataset based on the MNIST dataset.}
\label{fig:dataset_mnist}
\end{center}
\vskip -0.2in
\end{figure*}

We create a synthetic dataset using the MNIST dataset~\cite{lecun2010mnist}, which consists of 60,000 images for training and 10,000 images for testing. As described in Sec.~\ref{sec:exp_dataset_mnist}, each synthetic image consists of four digits, and its size becomes $56 \times 56$. Table~\ref{tab:dataset_mnist} presents the 12 class labels and the corresponding concepts. The order (positions) of the four digits in each synthetic image is randomly determined. When the synthetic images are generated, one of the four digits is dropped with a probability of 0.5. There are 68,017 synthetic images for training and 11,244 synthetic images for testing, which are generated with images in the official corresponding splits. Figure~\ref{fig:dataset_mnist} shows examples from the synthetic dataset.

For data augmentation, we use cutout~\cite{devries2017improved} with a size $7 \times 7$.

\begin{table}[h]
\caption{Pairs of class labels and concepts in the synthetic dataset based on the MNIST dataset.}
\vskip 0.15in
\begin{center}
\begin{small}
\begin{sc}
\begin{tabular}{cc}
\toprule
Class & Concepts \\
\midrule
0 & 0, 2, 4, 6 \\
1 & 0, 2, 5, 9 \\
2 & 0, 3, 4, 7 \\
3 & 0, 3, 6, 8 \\
4 & 1, 2, 5, 7 \\
5 & 1, 2, 4, 9 \\
6 & 1, 3, 5, 6 \\
7 & 1, 3, 7, 8 \\
8 & 1, 4, 6, 8 \\
9 & 3, 5, 7, 9 \\
10 & 2, 4, 7, 8 \\
11 & 2, 5, 6, 8 \\
\bottomrule
\end{tabular}\label{tab:dataset_mnist}
\end{sc}
\end{small}
\end{center}
\vskip -0.1in
\end{table}

\subsection{CUB}
During training, we use color jittering, horizontal flipping, and random scaling and cropping as data augmentation methods. Because strong random scaling and cropping can produce extreme noise in concept labels, we first resize the images to $256 \times 256$ and apply random scaling and cropping with a scale of $(0.8, 1.0)$. The final size of the images for training is $224 \times 224$. When we train the model with an image size of $299 \times 299$, the images are resized to $341 \times 341$ and cropped. For inference, the images are resized to $256 \times 256$ ($341 \times 341$) and center-cropped to $224 \times 224$ ($299 \times 299$).

Following \citet{koh2020concept}, we group concepts by using a common prefix of concept names in the dataset. The groups are used in concept intervention.

\subsection{AwA2}
Table~\ref{tab:awa2_attr} presents the remaining 45 attributes and how we group them for concept intervention. During training, we use the same data augmentation methods that are employed for the CUB dataset, except for resizing. For the AwA2 dataset, the images are not resized before cropping. For inference, the images are resized to $224 \times 224$.
\begin{table}[h]
\caption{The groups of attributes in AwA2 dataset.}
\vskip 0.15in
\begin{center}
\begin{small}
\begin{sc}
\begin{tabular}{lcccr}
\toprule
Group & attribute name \\
\midrule
color &black, white, blue, brown, gray, orange, red, yellow \\
pattern &patches, spots, stripes \\
hair &furry, hairless \\
skin &toughskin \\
size &big, small \\
fat &bulbous, lean \\
hand &flippers, hands, hooves, pads, paws \\
leg &long leg \\
neck &long neck \\
tail &tail \\
horn &horns \\
claw &claws \\
tusk &tusks \\
walk &bipedal, quadrapedal \\
\multirow{2}{*}{live} &arctic, coastal, desert, bush, plains, forest, fields, jungle, \\
& mountains, ocean, ground, water, tree, cave \\
\bottomrule
\end{tabular}\label{tab:awa2_attr}
\end{sc}
\end{small}
\end{center}
\vskip -0.1in
\end{table}

\section{Experimental details}\label{sec:appendix_exp_detail}
\subsection{Details for ProbCBM}
For the synthetic dataset, we set $D_c$ and $D_y$ as 16 and 32, respectively. For the real-world datasets, $D_c$ and $D_y$ are set as 16 and 128, respectively. $N_s$ is set as 50. With the synthetic dataset, we train the concept predictor for 30 epochs and the class predictor for 20 epochs. With the real-world datasets, we train the concept predictor for 50 epochs and the class predictor for 20 epochs. For stable training, we fix the pretrained weights with ImageNet-1K~\cite{russakovsky2015imagenet} for the first five epochs. We initialize $a$ and $d$, which are learnable parameters for scaling given by Eqs.~\ref{eq:prob_concept} and \ref{eq:prob_class_mc}, as 5 and 10, respectively.

We use AdamP optimizer~\cite{heo2021adamp} with the cosine learning rate scheduler~\cite{loshchilov2017sgdr}. The learning rate is set to $10^{-3}$ for the pretrained weights and $10^{-2}$ for the randomly initialized weights and learnable parameters. $\lambda_\text{KL}$ is set as $5\times 10^{-5}$. All experiments are implemented using PyTorch~\cite{paszke2017automatic}.

\subsection{Details for other models}
ResNet18 is used as a backbone in all models. For CBM, we incorporate an FC layer after the backbone to build a concept predictor. In addition, we utilize a separate FC layer as a class predictor. To apply MC dropout to CBM, we add a dropout layer with a dropout rate of 0.2 after every convolutional block in a backbone. During inference, the predictions were obtained with 50 samples same as ProbCBM. We train CBM and CBM with MC dropout in the same manner as our training scheme. The concept predictor outputs the probabilities that each concept exists, and the probabilities that the concept predictor generates are fed into the class predictor. We first train the concept predictor and then train the class predictor in a sequential way. During the training of the class predictor, the output of the concept predictor is replaced with a ground-truth concept label (1 and 0 for positive and negative labels, respectively) with probability $p_\text{replace}$. Equal to the case of training ProbCBM, $p_\text{replace}$ is set as 0.5.
CEM is trained with the official code\footnote{{\href{https://github.com/mateoespinosa/cem}{https://github.com/mateoespinosa/cem}}}. Please note that we use the same data augmentation techniques across all models while keeping other configurations at their default settings.

\section{Training scheme}
Algorithm \ref{alg:training_scheme} provides detailed steps of our training scheme for the class predictor, which is described in Sec.~\ref{sec:method_training}.
\begin{algorithm}[htb!]
   \caption{Training Scheme of Class Predictor}
   \label{alg:training_scheme}
\begin{algorithmic}
   \STATE {\bfseries Input:} Training data $\mathcal{D}$, Trained concept predictor
   \REPEAT
   \FOR{$x \in \mathcal{D}$}
   \FOR{concept $c \in \{c_1, c_2, ..., c_{N_c}\}$}
   \STATE Get $p(z_c|x)$ from concept predictor
   \FOR{$n=1$ {\bfseries to} $N_s$}
   \STATE Initialize representation set $\mathcal{Z}^{(n)}=\{\}$
   \STATE Sample $z_{c}^{(n)} \sim p(z_c|x)$
   \STATE Sample replace strategy $r \sim Bernoulli(p_\text{replace})$
   \IF{$r = true$}
   \STATE Append $\textbf{1}(c=1)z_c^+ + \textbf{1}(c=0)z_c^-$ to $\mathcal{Z}^{(n)}$
   \ELSE
   \STATE Append $z_{c}^{(n)}$ to $\mathcal{Z}^{(n)}$
   \ENDIF
   \ENDFOR
   \STATE Concatenate $\mathcal{Z}^{(n)}$ and get $h^{(n)}$ with Eq.~\ref{eq:proj_class}
   \ENDFOR
   \STATE Get $p(y_k = 1)$ with Eq.~\ref{eq:prob_class_mc}
   \STATE Calculate $\mathcal{L}_\text{class}$ and update the class predictor with $\mathcal{L}_\text{class}$
   \ENDFOR
   \UNTIL{the class predictor converges}
\end{algorithmic}
\end{algorithm}

\newpage
\section{Additional experimental results}

\subsection{Ablation studies}
We conduct ablations studies on the effect of the probability $p_\text{replace}$ and the embedding dimensions $D_c$ and $D_y$. The experiments are conducted with the CUB dataset.

\begin{figure}[h]
\begin{center}
\centerline{\includegraphics[width=0.28\columnwidth]{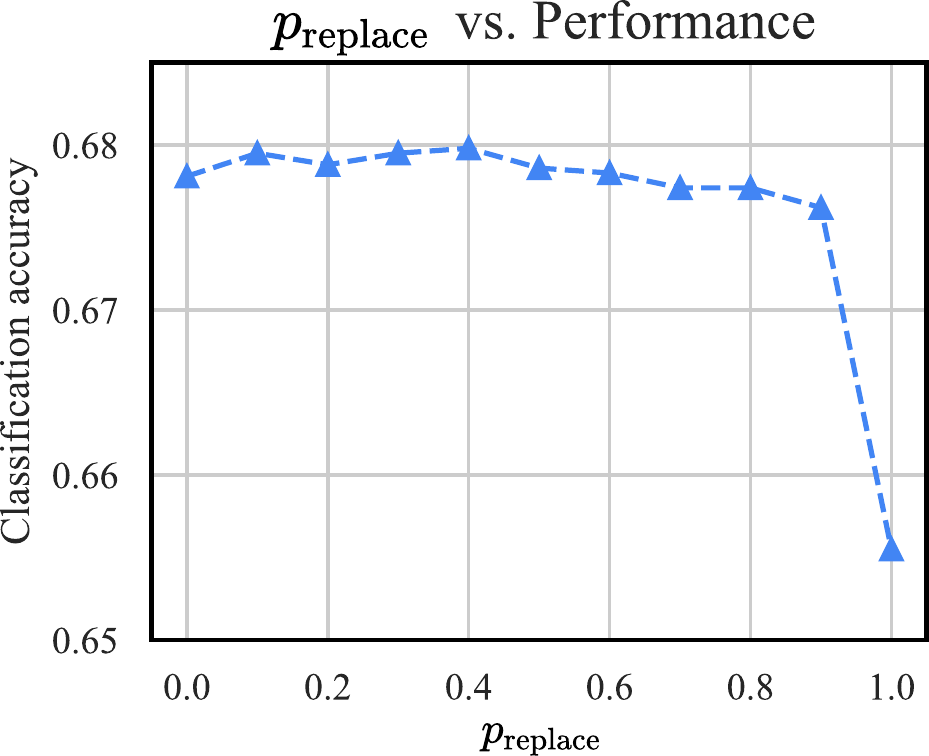}}
\vskip -0.12in
\caption{Classification performances of ProbCBM trained with different $p_\text{replace}$ for the CUB dataset.}
\label{fig:prob_replace}
\end{center}
\vskip -0.2in
\end{figure}

\textbf{Probability $p_\text{replace}$.}
Figure~\ref{fig:prob_replace} shows the classification performance with respect to $p_\text{replace}$. As shown, there is no significant difference except when $p_\text{replace}$ is 1. Training with $p_\text{replace}=1$ is the same as the independent learning strategy in CBM, wherein the class predictor is trained with ground-truth concept and class labels regardless of the concept predicted by the concept predictor.

\begin{figure}[htb!]
\begin{center}
\centerline{\includegraphics[width=0.62\columnwidth]{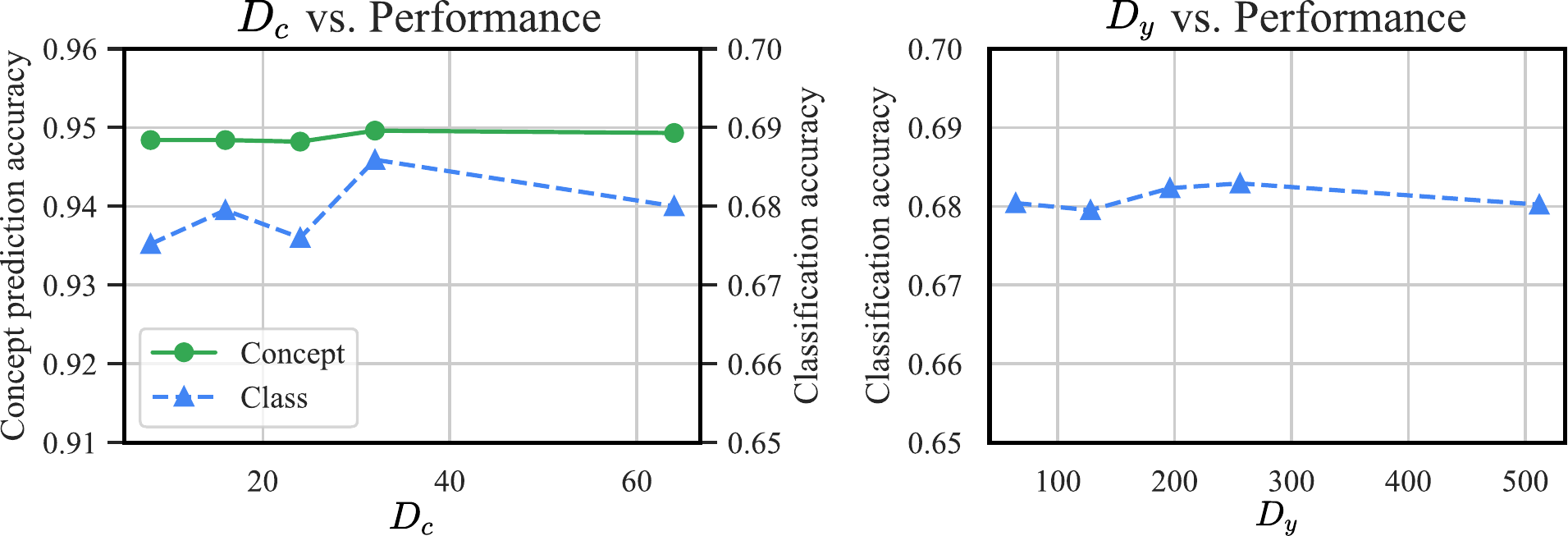}}
\vskip -0.12in
\caption{Concept prediction and classification performance of ProbCBM trained with different $D_c$ and $D_y$ values for the CUB dataset.}
\label{fig:ablation_d}
\end{center}
\vskip -0.2in
\end{figure}

\begin{figure}[htb!]
\begin{center}
\centerline{\includegraphics[width=0.38\columnwidth]{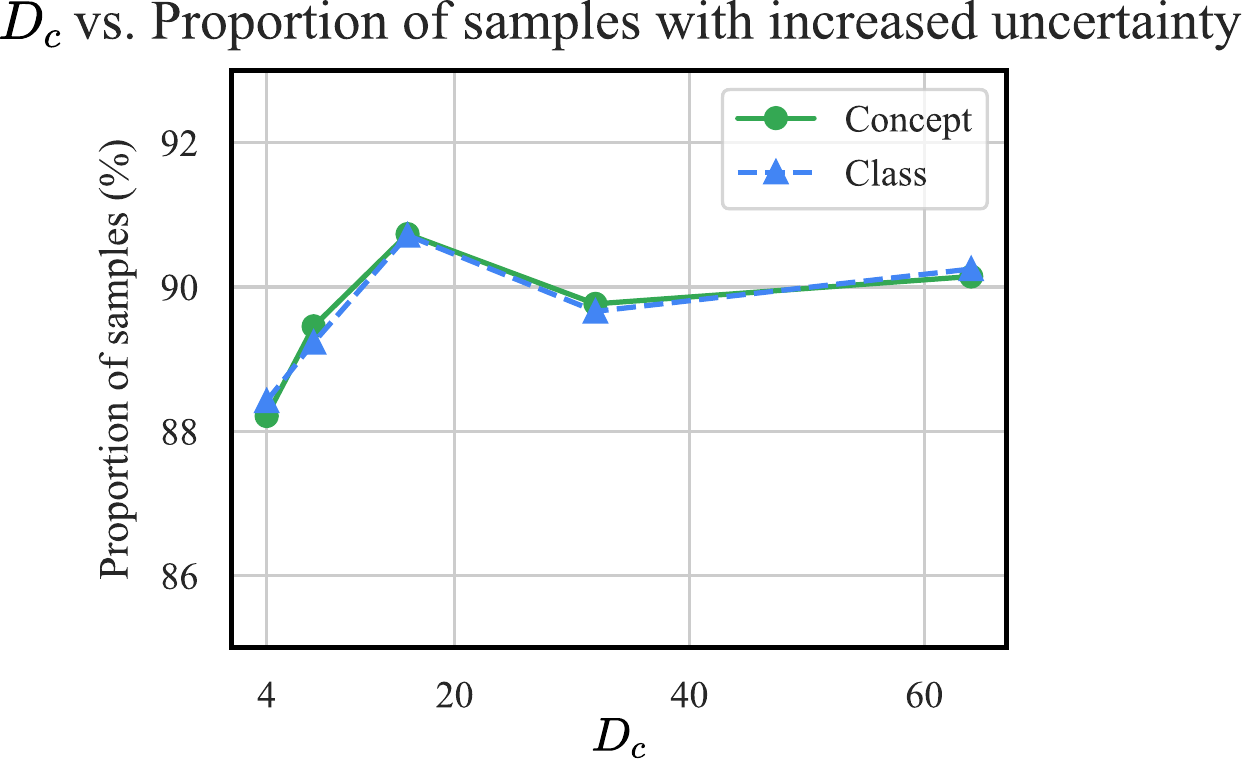}}
\vskip -0.12in
\caption{Proportion (\%) of samples whose uncertainty increases after occlusion of ProbCBM trained with different $D_c$ values for the CUB dataset.}
\label{fig:ablation_d_uncertainty}
\end{center}
\vskip -0.2in
\end{figure}

\textbf{Embedding dimensions $D_c$ and $D_y$.}
Figure~\ref{fig:ablation_d} shows the concept prediction and classification performance with respect to the embedding dimensions $D_c$ and $D_y$. $D_c$ is the dimension of the concept embedding space and $D_y$ is the dimension of the class embedding space. The concept prediction performance is the same in the ablation study on $D_y$. The performance is not significantly affected by $D_c$ or $D_y$. Additionally, we compare the proportions of samples that exhibit increased uncertainty after occlusion across different values of $D_c$, as depicted in Figure~\ref{fig:ablation_d_uncertainty}. The results indicate that when the concept embedding dimension $D_c$ is small ($D_c=4$), the ability to detect increased uncertainty diminishes. However, as $D_c$ increases over a certain threshold, it shows convergence. This finding indirectly suggests that utilizing high-dimensional concept embeddings has a positive impact on enhancing the ability for uncertainty estimation.


\subsection{Additional results}
\begin{figure}[h]
\begin{center}
\centerline{\includegraphics[width=0.55\columnwidth]{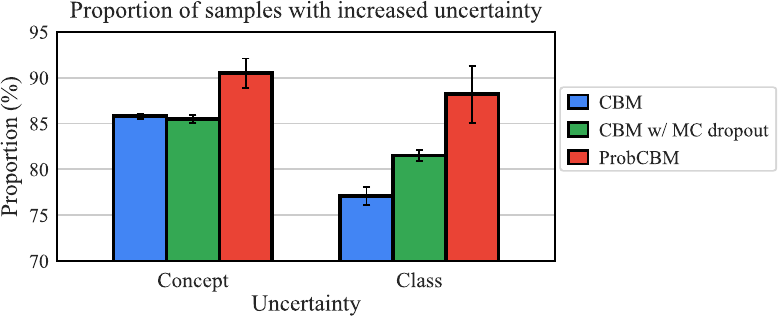}}
\vskip -0.05in
\caption{Proportion (\%) of samples whose uncertainty increases after occlusion for the AwA2 dataset. Results include mean values with standard deviation from three experiment repetitions.}
\label{fig:appendix_plot_uncertainty_inc_awa2}
\end{center}
\vskip -0.1in
\end{figure}

Figure~\ref{fig:appendix_plot_uncertainty_inc_awa2} presents the results of occlusion experiments conducted on the AwA2 dataset, similar to those performed on the CUB dataset in Section~\ref{sec:exp_cls_result}. We generate occlusion by covering a patch of size $64\times 64$ with gray color at the center of the images. The figure illustrates the proportion of samples that exhibit increased uncertainty after occlusion. The results demonstrate the effectiveness of ProbCBM in detecting an increase in both concept and class uncertainties across a diverse set of samples.

\begin{figure}[h]
\begin{center}
\centerline{\includegraphics[width=0.7\columnwidth]{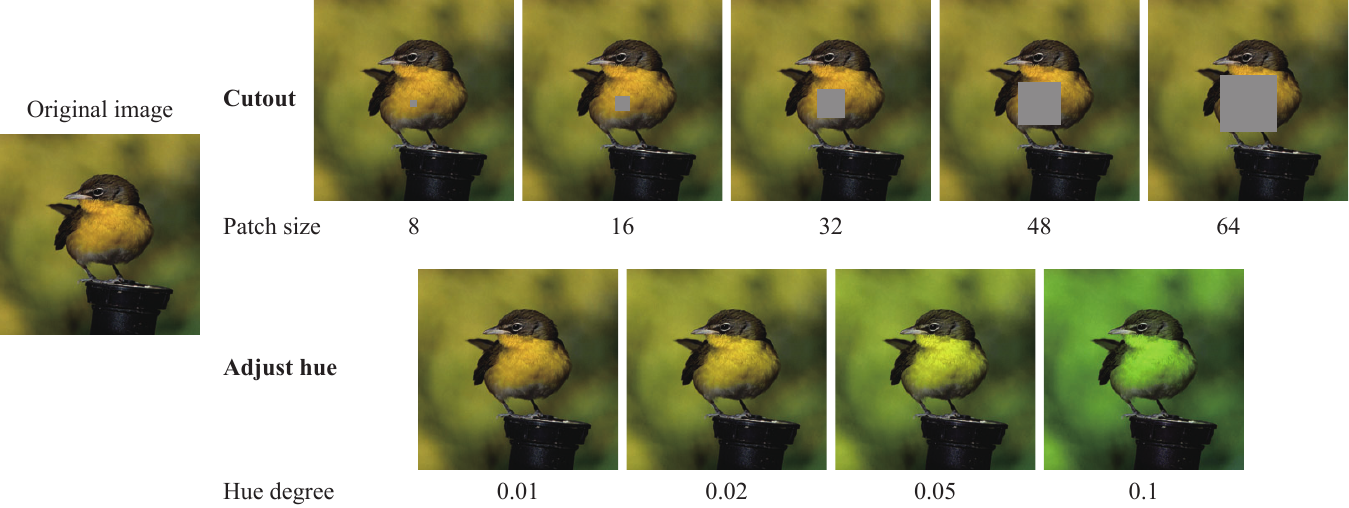}}
\vskip -0.05in
\caption{Examples of the resulting images after image transformations.}
\label{fig:appendix_cutout_example}
\end{center}
\vskip -0.2in
\end{figure}

\begin{figure*}[h]
\begin{center}
\centerline{\includegraphics[width=\textwidth]{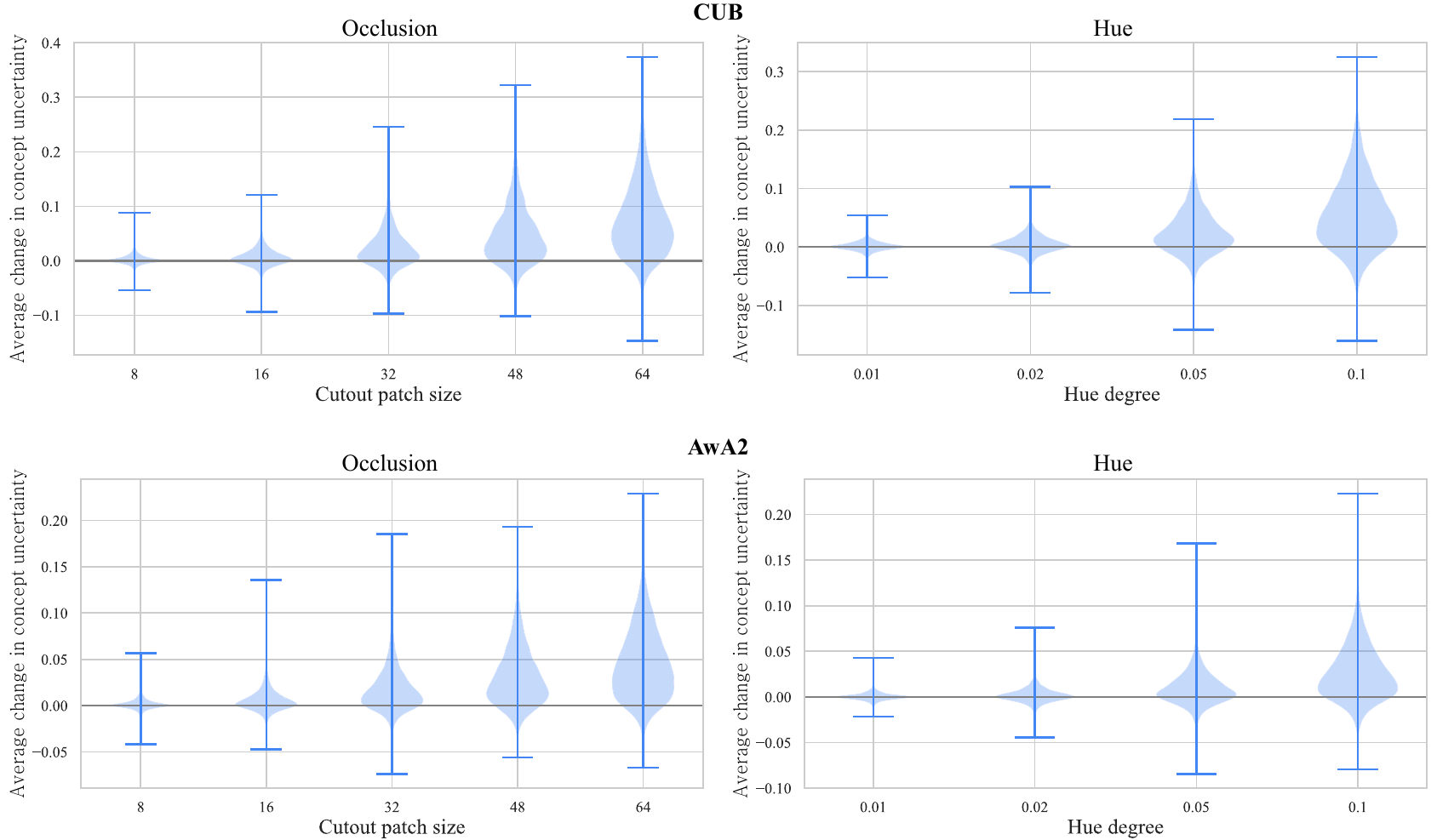}}
\vskip -0.05in
\caption{Violin plots of changes in the concept uncertainty after image transformations.}
\label{fig:cutout_violin}
\end{center}
\vskip -0.2in
\end{figure*}

Figure~\ref{fig:appendix_cutout_example} shows examples of the transformed images for the analysis presented in Sec.~\ref{sec:exp_real_uncertainty}. Figure~\ref{fig:cutout_violin} presents the changes in concept uncertainty after image transformations. As shown, the concept uncertainty increases for most samples. As the degree of image transformation increases, the proportion of samples with significantly increased uncertainty increases.

\begin{figure}[h]
\begin{center}
\centerline{\includegraphics[width=0.6\columnwidth]{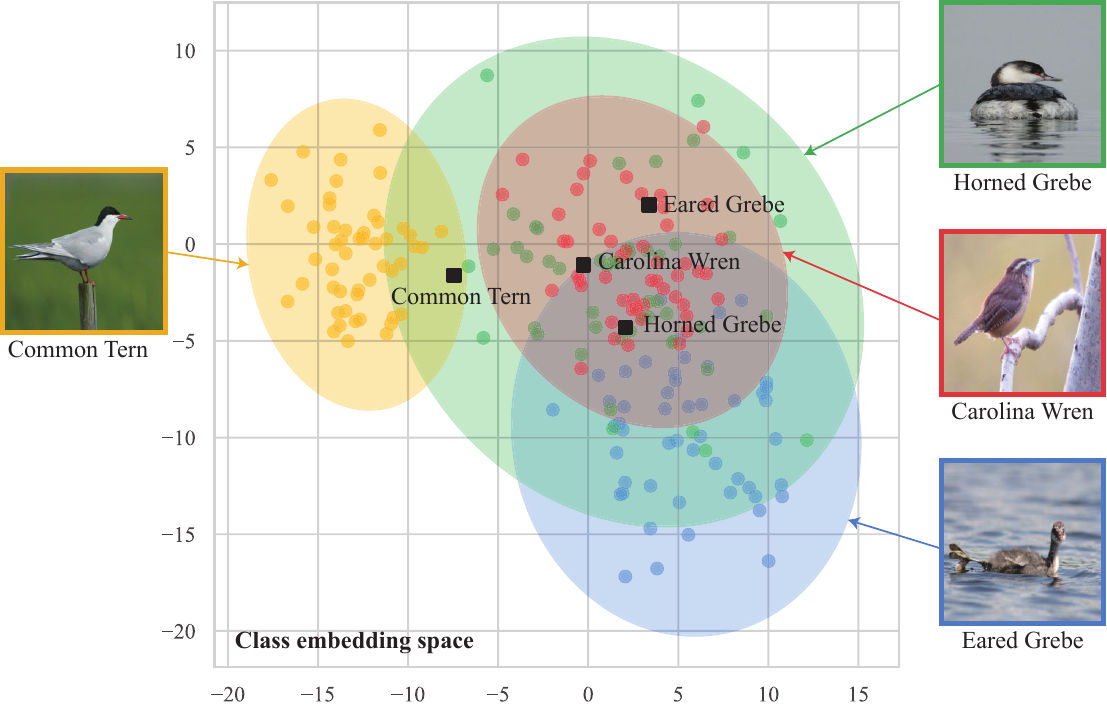}}
\vskip -0.05in
\caption{Visualization of the class embedding space.}
\label{fig:appendix_embedding}
\end{center}
\vskip -0.2in
\end{figure}

As shown in Figure~\ref{fig:appendix_embedding}, we visualize class embeddings in the 2D space, with the four images presented in Figure~\ref{fig:concept_embedding}. The images with large concept uncertainties have large uncertainties in classification.

\begin{figure}[h]
\begin{center}
\centerline{\includegraphics[width=0.6\columnwidth]{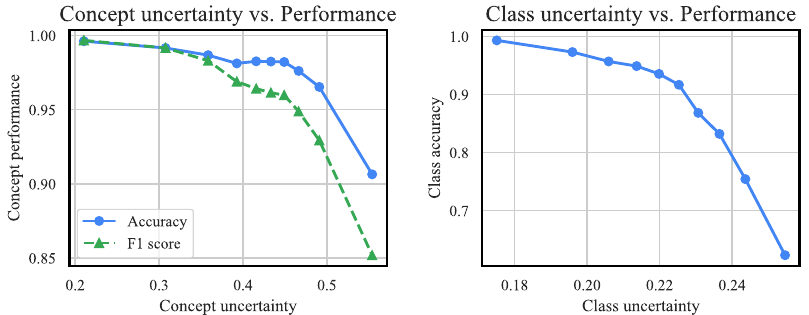}}
\vskip -0.05in
\caption{Correlation between the performance and uncertainty for the AwA2 dataset. A point represents one group of samples with similar uncertainties, and the x-axis indicates the median uncertainty for each group.}
\label{fig:perf_and_uncertainty_awa2}
\end{center}
\vskip -0.2in
\end{figure}

\begin{figure}[h]
\begin{center}
\centerline{\includegraphics[width=0.6\columnwidth]{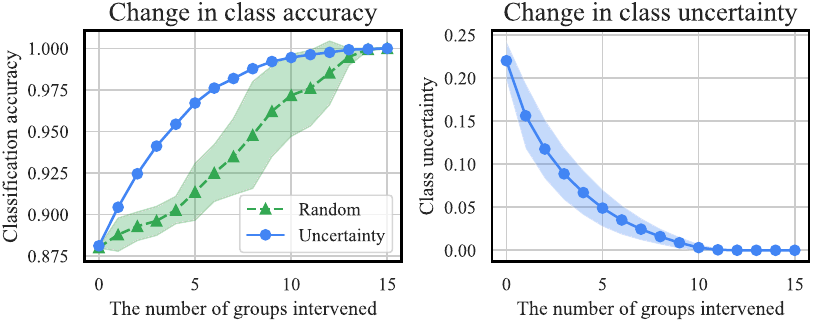}}
\vskip -0.05in
\caption{Results of concept intervention for the AwA2 dataset. Intervention in a random order is conducted five times and is visualized with the means and standard deviations. On the right plots, intervention is conducted in an uncertainty-based order, and the means and standard deviations of the class uncertainty for all test samples are shown.}
\label{fig:appendix_intervention_awa2}
\end{center}
\vskip -0.2in
\end{figure}

Figure~\ref{fig:perf_and_uncertainty_awa2} shows the correlation between the performance and the estimated uncertainty for the AwA2 dataset. Figure~\ref{fig:appendix_intervention_awa2} shows the results of concept intervention for the AwA2 dataset. As described in Sec.~\ref{sec:concept_intervention}, we intervene in the concepts in the same group at once (15 groups for the AwA2 dataset).

\subsection{Additional examples}

\begin{figure*}[t]
\begin{center}
\centerline{\includegraphics[width=\textwidth]{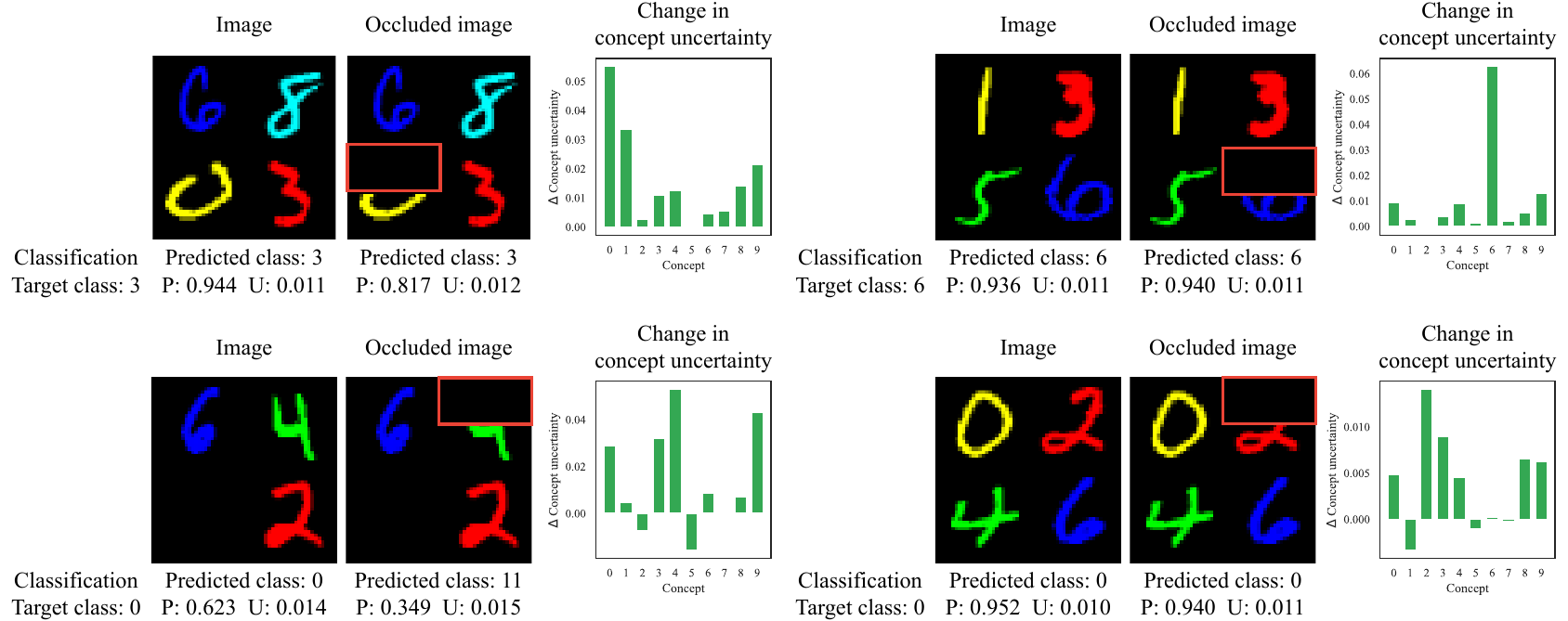}}
\vskip -0.05in
\caption{Examples of changes in the concept and class uncertainties after occlusion of parts of images. P and U present the probability and uncertainty, respectively. Red bounding boxes denote the occluded parts.}
\label{fig:appendix_example_mnist}
\end{center}
\vskip -0.2in
\end{figure*}

\begin{figure*}[t]
\begin{center}
\centerline{\includegraphics[width=\textwidth]{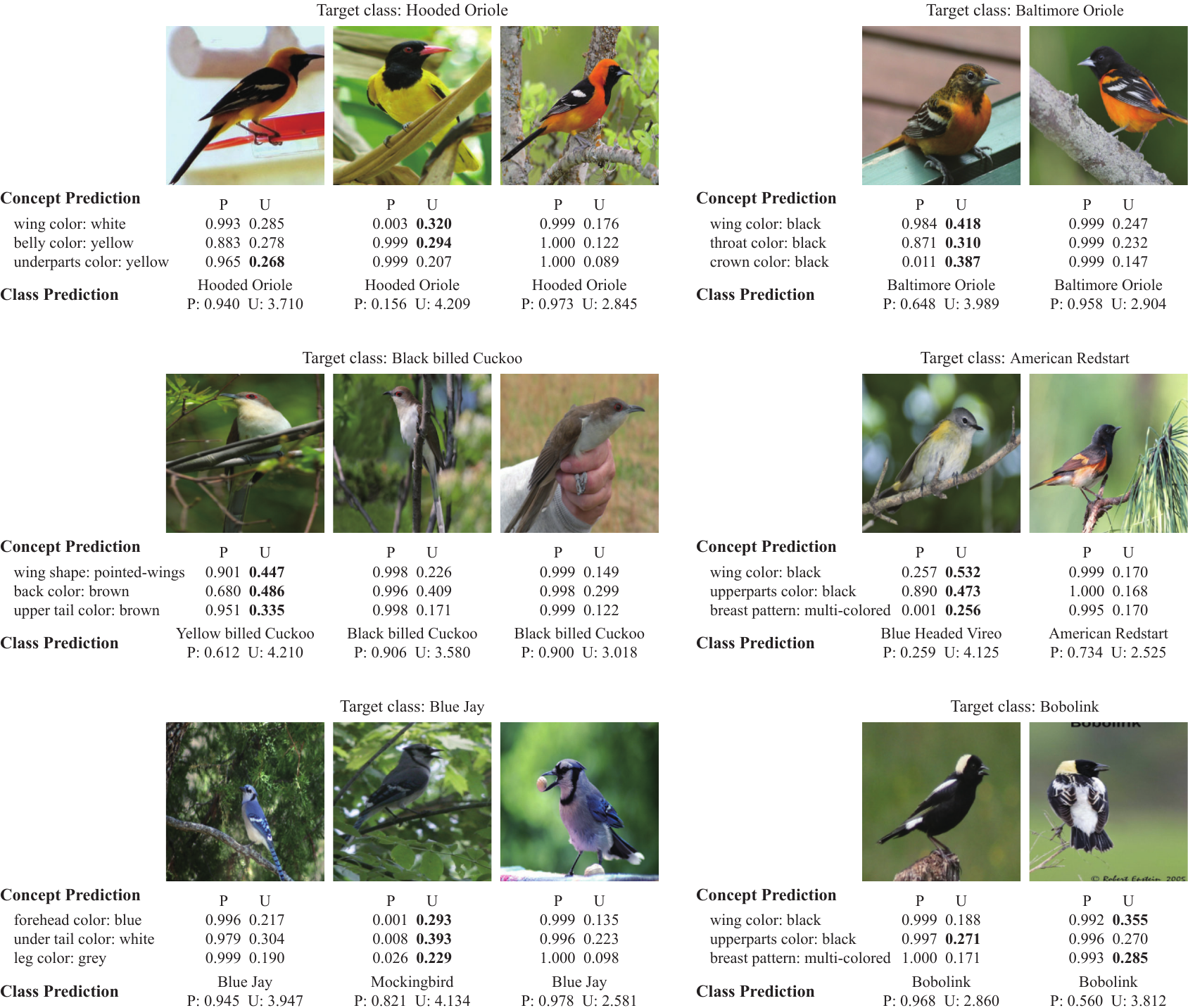}}
\caption{Examples of predictions of ProbCBM on the CUB dataset. P and U denote probability and uncertainty, respectively. The highest value of uncertainty in each concept is bolded.}
\label{fig:appendix_example_cub}
\end{center}
\vskip -0.2in
\end{figure*}

\begin{figure*}[t]
\begin{center}
\centerline{\includegraphics[width=\textwidth]{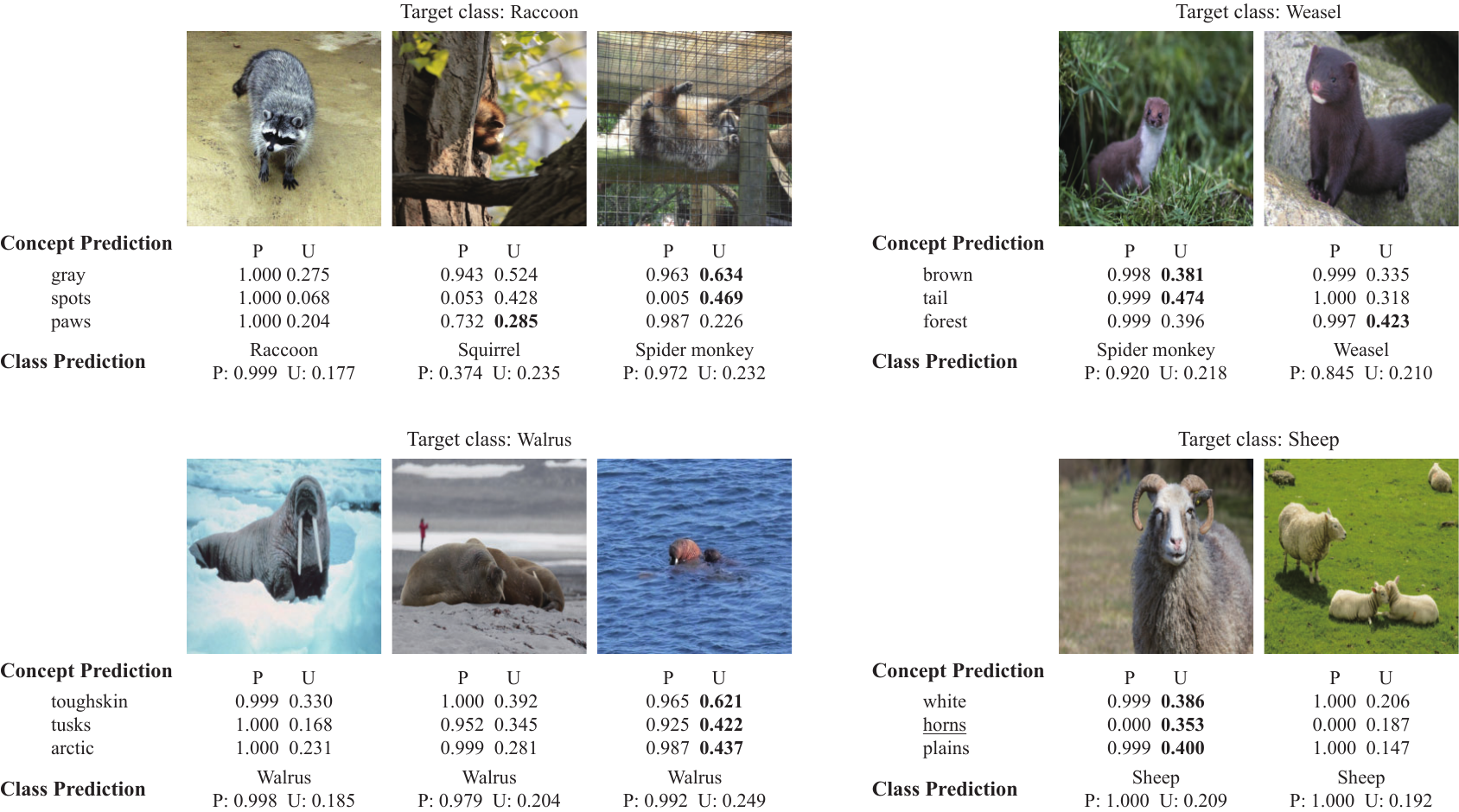}}
\caption{Examples of predictions of ProbCBM on the AwA2 dataset. P and U denote probability and uncertainty, respectively. The highest value of uncertainty in each concept is bolded. The labels of the most of concepts are positive except for underlined concepts.}
\label{fig:appendix_example_awa2}
\end{center}
\vskip -0.2in
\end{figure*}

Figure~\ref{fig:appendix_example_mnist} shows additional examples of changes in the predicted concept uncertainty of ProbCBM after occlusion of parts of images for the synthetic dataset. Figures~\ref{fig:appendix_example_cub} and \ref{fig:appendix_example_awa2} show additional examples of ProbCBM's prediction for the real-world datasets.

\vspace{20pt}


%%%%%%%%%%%%%%%%%%%%%%%%%%%%%%%%%%%%%%%%%%%%%%%%%%%%%%%%%%%%%%%%%%%%%%%%%%%%%%%
%%%%%%%%%%%%%%%%%%%%%%%%%%%%%%%%%%%%%%%%%%%%%%%%%%%%%%%%%%%%%%%%%%%%%%%%%%%%%%%

\end{document}